%% file: main.tex
\documentclass{article}
\usepackage{arxiv}
\input{preamble}

\renewcommand{\headeright}{Preprint}
\renewcommand{\undertitle}{Preprint}
\renewcommand{\shorttitle}{AnachroBench: Evaluating Historical Consistency in LVLMs}

\title{AnachroBench: Evaluating Historical Consistency in Large Vision-Language Models}
\date{}
\author{
  Keunhyeung Park$^{1}$\thanks{Equal contribution.}\quad
  Byeonggeuk Lim$^{2}$\footnotemark[1]\quad
  Kahyeon Kim$^{1}$\quad
  Sunhee Jo$^{2}$\\
  \bfseries Hayeong Ryu$^{1}$\quad
  Hoejoon Kwon$^{1}$\quad
  Jinhee Jang$^{1}$\quad
  YoungBin Kim$^{1,2}$\thanks{Corresponding author.}\\[6pt]
  \mdseries\small
  $^{1}$Department of Artificial Intelligence, Chung-Ang University\\
  \mdseries\small $^{2}$Graduate School of Advanced Imaging Science, Multimedia \& Film, Chung-Ang University\\[3pt]
  \small\{synoark99, banggeuk, kahyun0817, jo3438, bluebarry37, chriskwon9, jinheejang, ybkim85\}@cau.ac.kr
}

\begin{document}
\maketitle
\input{sec/00_Abstract}
\input{sec/01_Introduction}
\input{sec/02_Related_Work}
\input{sec/03_Method}
\input{sec/04_Experiments}
\FloatBarrier
\input{sec/05_Conclusion}
\bibliographystyle{splncs04}
\bibliography{main}

\appendix
\input{sec/06_Appendix}

\end{document}

%% file: preamble.tex
\usepackage[T1]{fontenc}
\usepackage{newtxtext}
\usepackage[varvw]{newtxmath}
\usepackage{graphicx}
\usepackage{xcolor}
\usepackage{microtype}
\usepackage{booktabs}
\usepackage{multirow}
\usepackage{makecell}
\usepackage{placeins}
\usepackage{adjustbox}
\usepackage{tcolorbox}
\tcbuselibrary{breakable}
\usepackage{pifont}
\usepackage[caption=false]{subfig}
\usepackage[hidelinks]{hyperref}
\newcommand{\best}[1]{\textbf{#1}}
\newcommand{\second}[1]{\underline{#1}}
\newcommand{\cmark}{\ding{51}}
\newcommand{\xmark}{\ding{55}}

\AtBeginEnvironment{itemize}{\setlength{\leftmargini}{1em}}

\newlength{\capinset}
\makeatletter
\patchcmd{\@makecaption}{#1: #2\par}{\leftskip\capinset \rightskip\capinset #1: #2\par}{}%
  {\PackageWarningNoLine{preamble}{caption inset patch 1 not applied}}
\patchcmd{\@makecaption}{\ifdim \wd\@tempboxa >\hsize}{\ifdim \wd\@tempboxa >\dimexpr\hsize-2\capinset\relax}{}%
  {\PackageWarningNoLine{preamble}{caption inset patch 2 not applied}}
\makeatother

%% file: sec/00_Abstract.tex
\begin{abstract}
Existing Large Vision-Language Model benchmarks primarily evaluate visual understanding in general everyday scenarios, leaving historical consistency with a specified era and country largely underexplored. To address this gap, we propose \textbf{AnachroBench}, a controlled discriminative benchmark for evaluating historical consistency in visual scenes. It contains 1,600 image pairs from eight countries spanning the late 19th to the early 20th century, each pairing an original historical image with an anachronistic version containing a seamlessly inserted modern object. AnachroBench evaluates three aspects: \textit{Inserted Object Perception}, which tests whether models detect the inserted modern object; \textit{Era-Country Consistency}, which assesses whether the scene is consistent with the specified era and country; and \textit{Conditional Era-Country Consistency}, which measures consistency judgments only when the inserted object is correctly perceived. A false consistency judgment occurs when a model perceives the modern object but still judges the scene as historically consistent. Overall, AnachroBench reveals a gap between inserted object perception and era-country consistency judgments, providing a targeted evaluation of this behavior.
\end{abstract}


%% file: sec/01_Introduction.tex
\section{Introduction}
\label{sec:introduction}
\input{figures/figure_01}

\looseness=-1
Large Language Models (LLMs)~\cite{openai2023gpt4} integrated with visual encoders have led to Large Vision-Language Models (LVLMs), demonstrating strong performance in visual question answering and image captioning~\cite{bai2025qwen25vl,gemini2025gemini25,liu2024llava15,liu2023llava,wang2025internvl35,ye2024mplugowl2}. Despite these advances, models may accurately perceive individual visual elements while failing to determine their compatibility with the broader temporal or geographic context, producing plausible but inconsistent responses~\cite{liu2024hallucinationsurvey,ji2023hallucination,li2023pope,rohrbach2018chair}.

LVLM evaluation has evolved from object existence~\cite{li2023pope,rohrbach2018chair} to object attributes and relations, visual illusions~\cite{chen2024rope,guan2024hallusionbench,kaul2024throne,tu2025ode}, and commonsense consistency~\cite{bittonguetta2023whoops,he2025rohe,liu2025phd}. However, existing benchmarks predominantly address everyday scenarios independent of specific eras or locations. Century~\cite{akbulut2025century} and TAB-VLM~\cite{ranjan2026tabvlm} evaluate historical contextualization and period reasoning on original images, but do not directly evaluate whether a scene is consistent with an explicitly specified era and country.

Visual consistency depends on historical context: even within the same country, objects natural in the 1900s differ from those of today. Assessing consistency therefore requires recognizing visual entities and reconciling them with the provided context. When LVLMs screen or curate historical image collections, such as AI-generated period imagery~\cite{palmini2026histvis} or mislabeled historical photographs, their consistency judgments are reliable only if they follow from the evidence the model itself perceives.

To examine how models use visual evidence in historical-context judgments, we propose \textbf{AnachroBench}, a controlled discriminative benchmark. Figure~\ref{fig:figure1} illustrates a model identifying a drone yet accepting the edited scene as consistent with France during 1870--1919. We call this a \emph{false consistency judgment}: an edited scene is accepted as consistent even after the inserted object has been perceived. This setting tests how perceived object evidence informs context-dependent decisions rather than relying on spontaneous free-form generation.

We collected historical images from eight countries spanning the late 19th to early 20th century and inserted modern objects to create anachronistic counterparts. Human validation of insertion position, size, and appearance yielded 1,600 original--anachronistic pairs. Finally, we evaluate three aspects: \textit{Inserted Object Perception}, \textit{Era-Country Consistency}, and \textit{Conditional Era-Country Consistency}.

Our contributions are: (1) AnachroBench, a controlled benchmark of historical consistency across eight countries; (2) evaluation of 14 open-source and four closed-source LVLMs on the three metrics; and (3) analyses of perception--consistency gaps, examining evidence use and the effect of explicit object-timing information.

%% file: figures/figure_01.tex
\begin{figure}[t]
    \centering
    \includegraphics[width=0.95\textwidth]{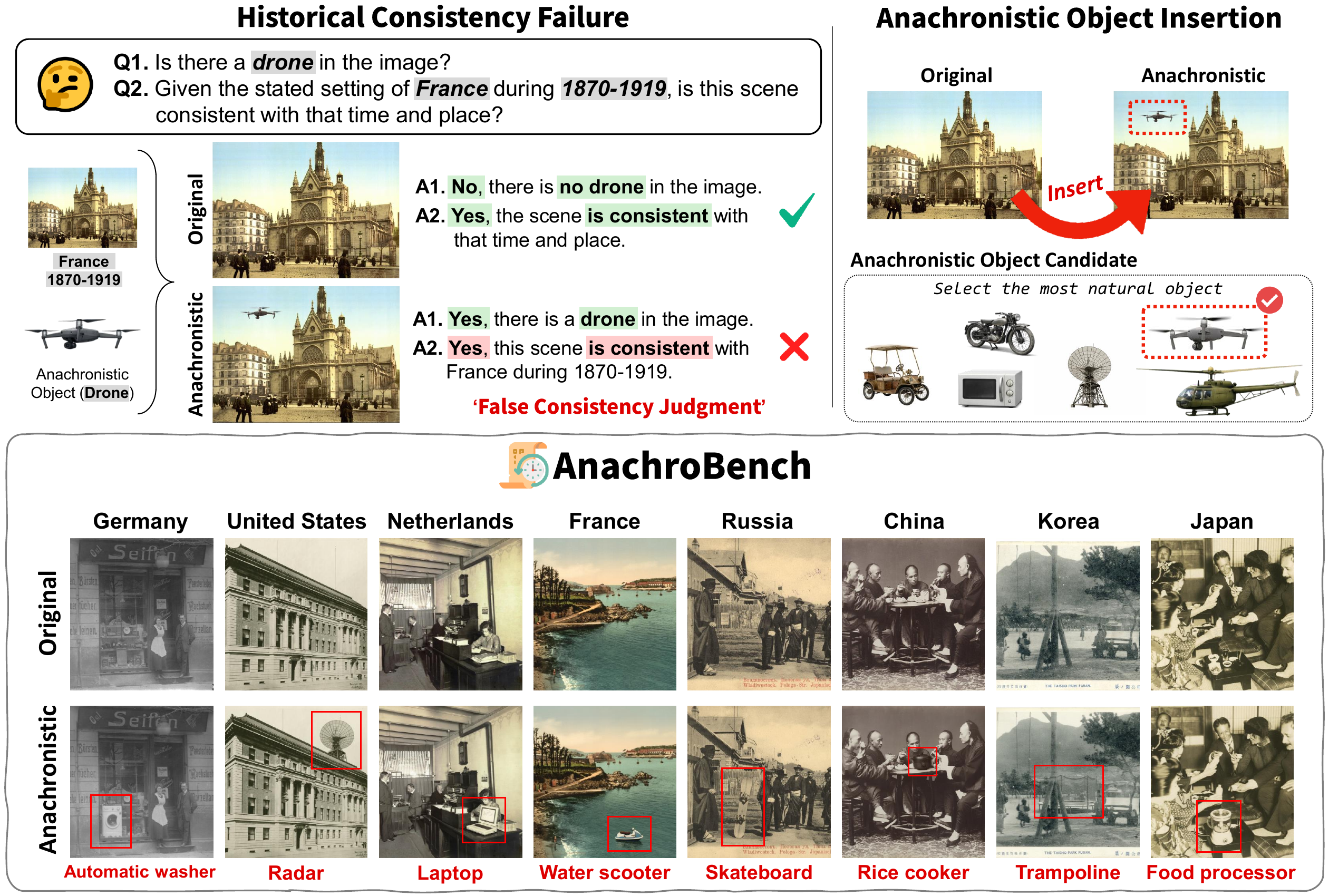}
    \caption{Overview of false consistency judgments and the AnachroBench benchmark. \textbf{(Top-left)} Example of a false consistency judgment: the model perceives the inserted drone but still judges the edited scene to be consistent with the historical setting. \textbf{(Top-right)} Illustration of Anachronistic Object Insertion. \textbf{(Bottom)} Representative image pairs from AnachroBench across eight countries. Each pair consists of an original image (top row) and an anachronistic counterpart (bottom row), with inserted objects highlighted in \textcolor{red}{red}.}
    \label{fig:figure1}
\end{figure}

%% file: sec/02_Related_Work.tex
\section{Related Work}
\label{sec:related_work}
\subsection{Large Vision-Language Models}
\label{sec:lvlm}
\input{tables/table_01}

\looseness=-1
The rapid evolution of LLMs has driven the development of LVLMs, which integrate visual encoders with language decoders to enable strong multimodal instruction-following capabilities. Open-source LVLMs include LLaVA~\cite{liu2023llava}, InstructBLIP~\cite{dai2023instructblip}, mPLUG-Owl2~\cite{ye2024mplugowl2}, CogVLM2~\cite{hong2024cogvlm2}, InternVL3.5~\cite{wang2025internvl35}, Gemma 3~\cite{kamath2025gemma3}, and Llama-3.2-Vision~\cite{dubey2024llama3}; closed-source systems include GPT-4o~\cite{openai2024gpt4o}, Gemini 2.5~\cite{gemini2025gemini25}, and Claude Sonnet~\cite{anthropic2024claude}. Despite this progress, LVLMs may still produce responses inconsistent with visual evidence or fail to incorporate perceived evidence into contextual judgments.

\subsection{LVLM Evaluation Benchmarks} 
Early LVLM benchmarks such as CHAIR~\cite{rohrbach2018chair} and POPE~\cite{li2023pope} focused on verifying the presence of isolated objects. HallusionBench~\cite{guan2024hallusionbench} tests answer consistency on original--edited pairs, ROPE~\cite{chen2024rope} and THRONE~\cite{kaul2024throne} evaluate multi-object and free-form object hallucination, and ROHE~\cite{he2025rohe}, PhD~\cite{liu2025phd}, and ODE~\cite{tu2025ode} add object removal, prompted questions, and open-set evaluation. COinCO~\cite{yang2026coinco} inpaints out-of-context objects, whereas its context is defined by location, size, and co-occurrence in everyday scenes. Century~\cite{akbulut2025century} evaluates free-form historical contextualization of geographically diverse images, TAB-VLM~\cite{ranjan2026tabvlm} measures cultural anachronism through period questions on Indian artifacts, and HistVis~\cite{palmini2026histvis} examines anachronistic objects in text-to-image outputs. AnachroBench instead uses controlled object insertion into real photographs and discriminative evaluation to test whether correctly perceived evidence is incorporated into a judgment about an explicitly specified era and country (Table~\ref{tab:benchmark_comparison}).

%% file: tables/table_01.tex
\begin{table}[t]
\caption{Comparison with representative LVLM benchmarks. Context indicates the use of era/country information. Task types are abbreviated as Disc. for discriminative and Gen. for generative.}
\centering
\footnotesize
\setlength{\tabcolsep}{4pt}
\renewcommand{\arraystretch}{1.0}
\makebox[\linewidth][c]{%
\adjustbox{max width=0.95\linewidth}{%
\begin{tabular}{l r r l l c c c}
\toprule
\multirow{2}{*}{\raisebox{-0.5ex}{\textbf{Benchmark}}}
    & \multirow{2}{*}{\raisebox{-0.5ex}{\textbf{Images}}}
    & \multirow{2}{*}{\raisebox{-0.5ex}{\textbf{Eval Units}}}
    & \multirow{2}{*}{\raisebox{-0.5ex}{\textbf{Construction}}}
    & \multirow{2}{*}{\raisebox{-0.5ex}{\textbf{Format}}}
    & \multirow{2}{*}{%
        \raisebox{-0.5ex}[0pt][0pt]{%
            \makecell{\textbf{LLM-free}\\\textbf{Eval}}
        }%
      }
    & \multicolumn{2}{c}{\textbf{Context}} \\
\cmidrule(lr){7-8}
& & & & & & \textbf{Era} & \textbf{Country} \\
\midrule
POPE~\cite{li2023pope} & 500 & 3,000 & Original & Disc. & \textcolor{green!60!black}{\cmark} & \textcolor{red}{\xmark} & \textcolor{red}{\xmark} \\
AMBER~\cite{wang2024amber} & 1,004 & 15,220 & Original & Mixed & \textcolor{green!60!black}{\cmark} & \textcolor{red}{\xmark} & \textcolor{red}{\xmark} \\
THRONE~\cite{kaul2024throne} & 5,000 & 400,000 & Original & Gen. & \textcolor{red}{\xmark} & \textcolor{red}{\xmark} & \textcolor{red}{\xmark} \\
ROPE~\cite{chen2024rope} & 4,661 & 23,305 & Original & Disc. & \textcolor{green!60!black}{\cmark} & \textcolor{red}{\xmark} & \textcolor{red}{\xmark} \\
HallusionBench~\cite{guan2024hallusionbench} & 346 & 1,129 & Original + Edited & Disc. & \textcolor{red}{\xmark} & \textcolor{red}{\xmark} & \textcolor{red}{\xmark} \\
WHOOPS!~\cite{bittonguetta2023whoops} & 500 & 3,374 & Synthetic & Mixed & \textcolor{red}{\xmark} & \textcolor{red}{\xmark} & \textcolor{red}{\xmark} \\
ODE~\cite{tu2025ode} & 808 & 8,786 & Synthetic & Mixed & \textcolor{green!60!black}{\cmark} & \textcolor{red}{\xmark} & \textcolor{red}{\xmark} \\
ROHE~\cite{he2025rohe} & 11,008 & 11,008 & Original + Edited & Disc. & \textcolor{green!60!black}{\cmark} & \textcolor{red}{\xmark} & \textcolor{red}{\xmark} \\
PhD~\cite{liu2025phd} & 15,398 & 102,564 & Original + Synthetic & Disc. & \textcolor{green!60!black}{\cmark} & \textcolor{red}{\xmark} & \textcolor{red}{\xmark} \\
Century~\cite{akbulut2025century} & 1,500 & 1,500 & Original & Gen. & \textcolor{red}{\xmark} & \textcolor{green!60!black}{\cmark} & \textcolor{green!60!black}{\cmark} \\
TAB-VLM~\cite{ranjan2026tabvlm} & 1,600 & 600 & Original & Disc. & \textcolor{green!60!black}{\cmark} & \textcolor{green!60!black}{\cmark} & \textcolor{red}{\xmark} \\
\midrule
\textbf{AnachroBench (Ours)} & \textbf{3,200} & \textbf{6,400} & \textbf{Original + Synthetic} & \textbf{Disc.} & \textcolor{green!60!black}{\cmark} & \textcolor{green!60!black}{\cmark} & \textcolor{green!60!black}{\cmark} \\
\bottomrule
\end{tabular}
}
}
\label{tab:benchmark_comparison}
\end{table}

%% file: sec/03_Method.tex
\section{AnachroBench Evaluation Suite}
\label{sec:dataset_construction}
\label{sec:object_insertion}
\subsection{Benchmark Construction}
\label{sec:benchmark_construction}
We used three \href{https://deepmind.google/technologies/gemini/}{Gemini models}: \texttt{gemini-\allowbreak{}3-\allowbreak{}flash-\allowbreak{}preview} for filtering and object selection, \texttt{gemini-\allowbreak{}3.1-\allowbreak{}pro-\allowbreak{}preview} for anachronistic object classification, and \texttt{gemini-\allowbreak{}2.5-\allowbreak{}flash-\allowbreak{}image} for compositing (prompts \mbox{in Appendix~\ref{sec:appendix_prompt}).}

\subsubsection{Data Collection and Filtering}
\label{sec:data_collection}
We collected 8,456 historical photographs from \href{https://commons.wikimedia.org}{Wikimedia Commons}, the \href{https://digitalcollections.nypl.org}{New York Public Library (NYPL)}, the \href{https://www.loc.gov}{Library of Congress (LoC)}, and \href{https://gongu.copyright.or.kr}{GongU Madang}, a Korean public-domain archive (source breakdown in Appendix~\ref{sec:appendix_statistics}). Images cover Germany (DE), United States (US), Netherlands (NL), France (FR), Russia (RU), Korea (KR), China (CN), and Japan (JP) during 1870--1929, a period that predates the inserted objects and offers abundant public-domain photographs with reliable date and location metadata. We retained images whose licenses or rights statements permit modification and redistribution, including Public Domain, CC BY, CC BY-SA, and CC0. The image pairs, metadata, and evaluation prompts will be made publicly available.

\looseness=-1
Model-assisted filtering first removed violent, sexually explicit, or hateful content, then images lacking clear contextual cues indicative of their era. Of the 8,456 collected images, 3,489 (41.3\%) passed filtering and proceeded to object insertion.

\textbf{Object pool construction.}
We drew candidates from LVIS~\cite{gupta2019lvis} and Objects365~\cite{shao2019objects365}, covering 1,203 and 365 categories. A model-assisted procedure selected objects invented or first widely adopted after 1930. Authors removed candidates with ambiguous historical status or insufficiently distinctive appearance, yielding 45 anachronistic objects.

\textbf{Object--image matching and compositing.}
For each image, the model selected the most contextually appropriate anachronistic object and inserted it at a physically plausible location, preserving the photograph's appearance and style.

\subsubsection{Human Validation}
\label{sec:quality_control}
A shared pool of five annotators familiar with the relevant historical and national contexts reviewed composite quality and plausibility. Two distinct annotators reviewed each country's candidates; assignments could overlap across countries. Each image was reviewed by exactly two annotators and retained only if both accepted all criteria. Of 3,489 candidates, 1,600 were retained. Cohen's $\kappa$ was 0.781, with 89.1\% agreement (Appendix~\ref{sec:appendix_iaa}). Annotators checked historical incompatibility, placement, scale and perspective, color and photographic style, and visible compositing artifacts; failure on any criterion led to rejection. Disagreements mainly concerned object size and placement.

\subsection{Evaluation Metrics}
\label{sec:evaluation_metrics}
We evaluate \textit{Inserted Object Perception} (IOP), \textit{Era-Country Consistency} (ECC), and \textit{Conditional Era-Country Consistency} (cECC). Each question appends ``Please answer yes or no,'' supporting straightforward quantitative evaluation~\cite{fu2023mme,li2023pope}. For IOP/ECC, a pair is correct only if both original and edited images are answered correctly, preventing success by repeating the same answer.

\input{figures/figure_02}
\begin{itemize}
\item \textbf{IOP}: ``Is there a \{object\} in the image?'' The ground truth is \textit{No} for the original and \textit{Yes} for the anachronistic image.
\item \textbf{ECC}: ``Given the stated setting of \{country\} during \{period\}, is this scene consistent with that time and place?'' The ground truth is \textit{Yes} for the original and \textit{No} for the anachronistic image.
\item \textbf{cECC}: To evaluate consistency judgments after successful object perception, we compute the proportion of IOP-correct anachronistic images judged inconsistent with the setting. Its model-specific denominator counts edited images answered correctly on IOP. An incorrect ECC answer on them is a false consistency judgment.
\end{itemize}
ECC asks about the whole scene so that the model must select the relevant evidence. The inserted object provides the temporal inconsistency; country and period specify the historical setting. The edits manipulate only temporal compatibility; country is supplied as context, so ECC does not separately measure geographic consistency.

\subsection{Dataset Analysis}
\label{sec:data_statistics}
AnachroBench contains 1,600 pairs (3,200 images) across eight countries and six decades. Germany and the United States are the largest subsets, and the 1900s are most densely represented (Figure~\ref{fig:figure2a}). Wikimedia Commons and LoC are the primary sources.

The 45 inserted objects follow a long-tail distribution shaped by human validation, from 115 pairs (golf cart) to one (CD player). Country-specific distributions reflect source scenes, contextually plausible object matching, and human acceptance (Figure~\ref{fig:figure2b}).

%% file: figures/figure_02.tex
\begin{figure}[b]
    \centering
    \subfloat[]{\begin{minipage}[t]{0.47\textwidth}
        \centering
        \includegraphics[width=0.98\linewidth]{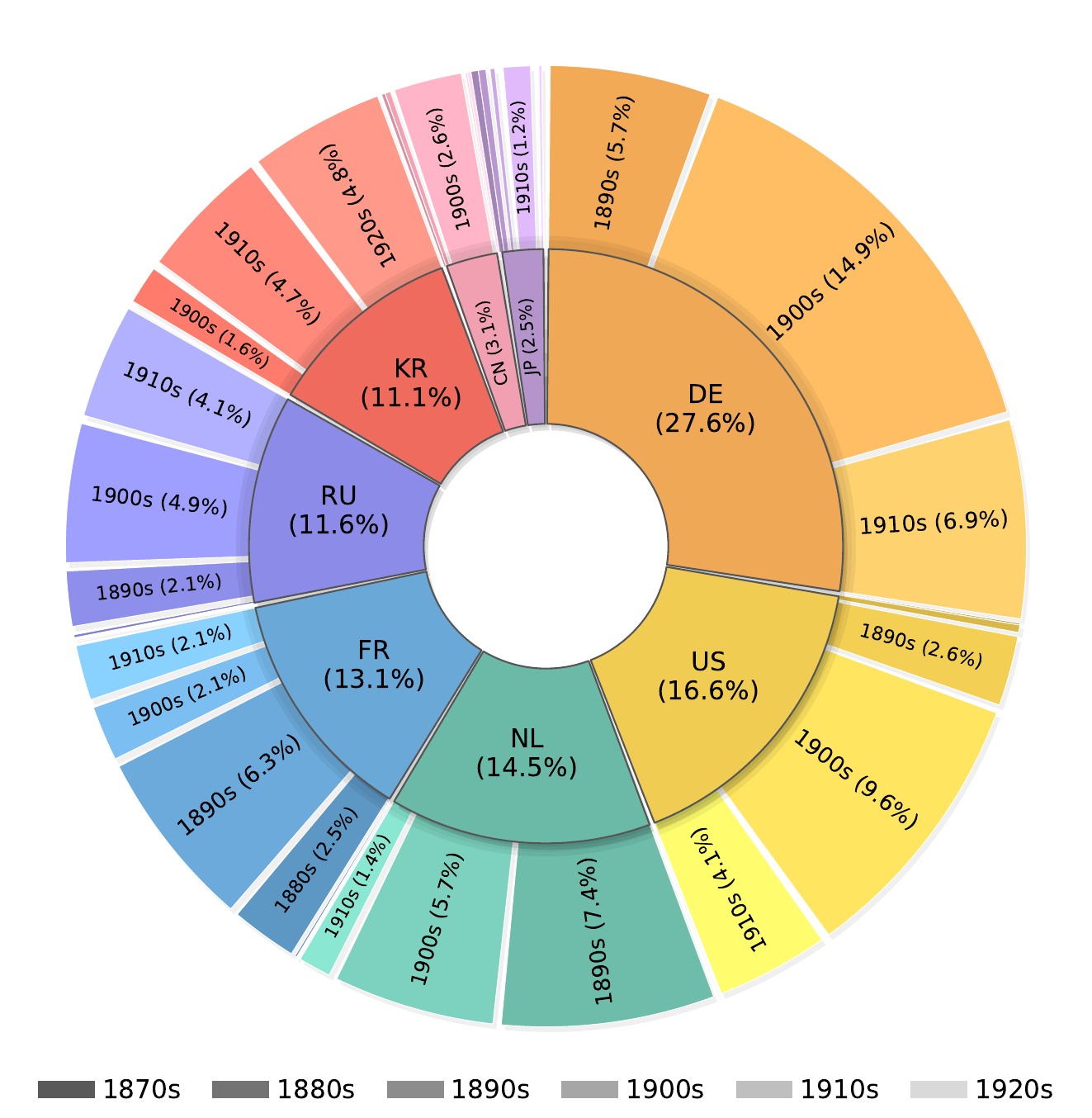}
        \label{fig:figure2a}
    \end{minipage}}
    \hfill
    \subfloat[]{\begin{minipage}[t]{0.47\textwidth}
        \centering
        \raisebox{3mm}{%
            \includegraphics[width=0.98\linewidth]{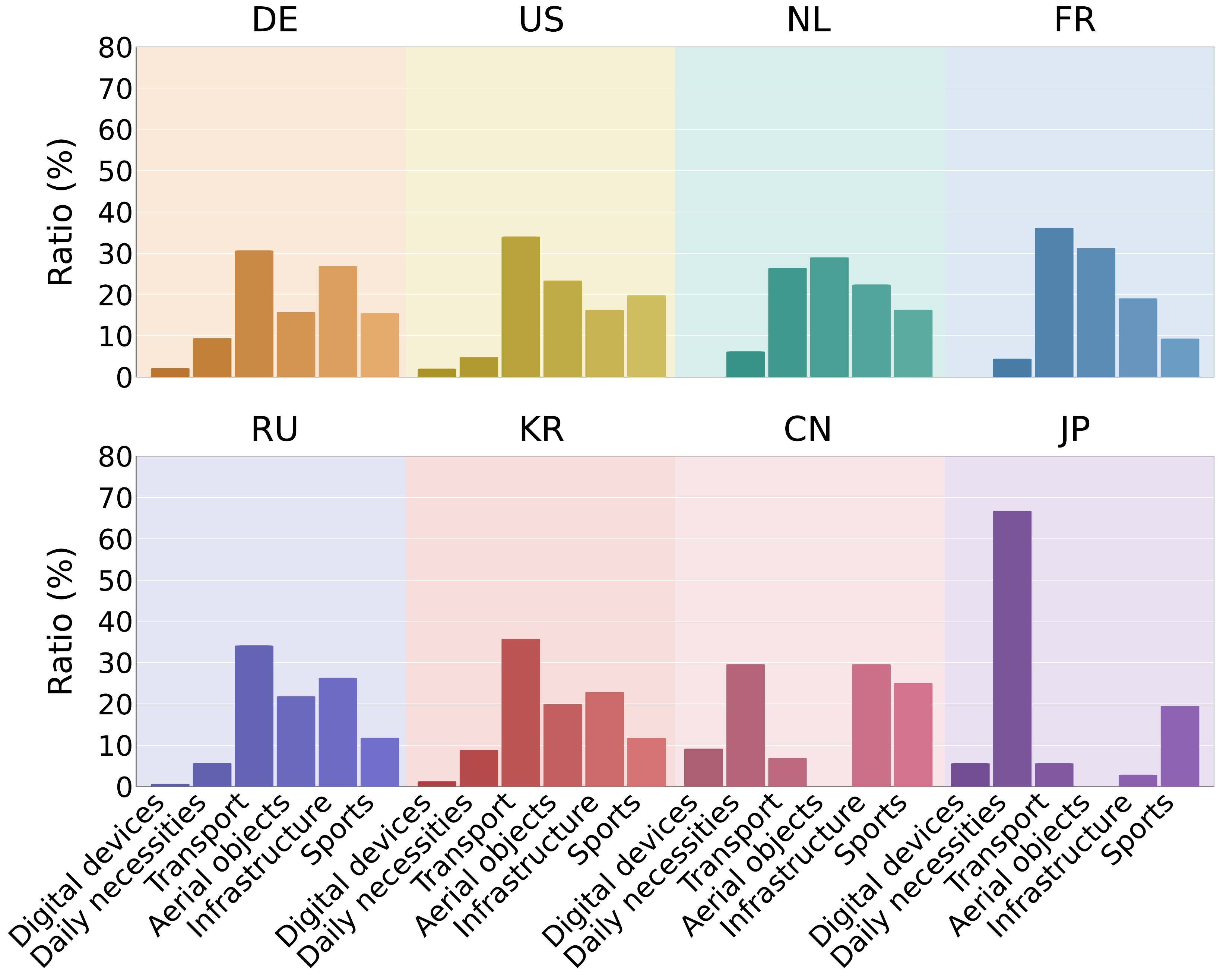}%
        }%
        \label{fig:figure2b}
    \end{minipage}}

    \caption{Dataset analysis. \textbf{(a)} Joint distribution of 1,600 benchmark pairs by country (inner ring) and decade (outer ring). \textbf{(b)} Per-country distribution of anachronistic objects grouped by category.}
    \label{fig:figure2}
\end{figure}

%% file: sec/04_Experiments.tex
\section{Experiments}
\label{sec:experiments}
\input{tables/table_02}

\subsection{Models}
\label{sec:setup}

We evaluated four closed-source models---GPT-4o~\cite{openai2024gpt4o}, Claude-Sonnet-4.5~\cite{anthropic2024claude}, and Gemini-2.5-Flash/Pro~\cite{gemini2025gemini25}---and 14 open-source models: InstructBLIP-7B/13B~\cite{dai2023instructblip}, mPLUG-Owl2-7B~\cite{ye2024mplugowl2}, LLaVA-1.5-7B/13B~\cite{liu2024llava15}, CogVLM2-19B~\cite{hong2024cogvlm2}, Llama-3.2-11B-Vision (Llama-3.2-11B)~\cite{dubey2024llama3}, Qwen2.5-VL-3B/7B~\cite{bai2025qwen25vl}, Gemma3-4B/12B/27B~\cite{kamath2025gemma3}, and InternVL3.5-8B/14B~\cite{wang2025internvl35}. As a human reference, three independent evaluators, unaffiliated with the benchmark construction process, each assessed the same 40 image pairs per country; Tables~\ref{tab:main1}--\ref{tab:main2} report their average performance.

\input{tables/table_03}

\subsection{Results}
\label{sec:results}

\noindent\textbf{Results on \textit{Inserted Object Perception}.}
Table~\ref{tab:main1} shows Llama-3.2-11B attaining the highest IOP in most countries. Open-source models perform comparably to closed-source models, and increasing model size within a family does not consistently improve IOP.

\noindent\textbf{Results on \textit{Era-Country Consistency}.}
ECC is markedly lower than IOP for most models (Table~\ref{tab:main1}; Appendices~\ref{sec:appendix_radar} and~\ref{sec:appendix_breakdowns}). Closed-source systems generally outperform open-source systems, with Gemini leading most countries. GPT-4o falls well below its own IOP; InstructBLIP is near zero; and Llama-3.2-11B remains weak despite its strong perception. Thus, recognizing an object does not ensure a correct judgment about its historical context.

\noindent\textbf{Results on Conditional \textit{Era-Country Consistency}.}
On IOP-correct edited images (Table~\ref{tab:main2}), both Gemini variants exceed 91\% overall cECC, whereas most open-source models remain below 50\%. False consistency judgments therefore persist after successful perception, when models must relate the perceived object to the historical setting.

\input{tables/table_04}

\subsection{Analysis}
\label{sec:analysis}

\input{figures/figure_03}

\noindent\textbf{Historically Consistent Control.}
To test whether models reject edited images because of the edit itself, we constructed 572 historically consistent images using the same pipeline and 23 era-appropriate objects across eight countries. Human validation achieved $\kappa=0.687$ (Appendices~\ref{sec:appendix_consistent} and~\ref{sec:appendix_examples}). Once the object is perceived, the four closed-source models accept 84.3--99.6\% of era-appropriate insertions (Table~\ref{tab:submain}) but reject 73.8--92.2\% of anachronistic ones (Table~\ref{tab:main2}), suggesting that rejection is not driven by the edit alone.

\noindent\textbf{Object-Level Variation.}
Figure~\ref{fig:figure4} reports cECC failure rates for the four closed-source models on the 20 most challenging objects. Difficulty varies substantially: all four perform poorly on Frisbee (32 benchmark pairs) and Pegboard (6). Averaged over all 18 models (Appendix~\ref{sec:appendix_breakdowns}), traffic cones reach 92.8\% IOP but 20.1\% cECC, whereas helicopters reach 96.2\% and 59.9\%.

\begin{table}[t]
\begin{minipage}[t]{0.35\linewidth}
\input{tables/table_05}
\end{minipage}
\hfill
\begin{minipage}[t]{0.59\linewidth}
\input{tables/table_06}
\end{minipage}
\end{table}

\looseness=-1
\noindent\textbf{Cross-Compositor Robustness.}
Replacing Gemini compositing with \texttt{gpt-\allowbreak{}image-\allowbreak{}1.5} yielded 689 regenerated pairs, of which both annotators accepted 432. Table~\ref{tab:main_gpt} reports six of the 11 rerun models (full results in Appendix~\ref{sec:appendix_gpt_compositing}). Gemini variants again rank highest among all 11, indicating that the performance pattern persists with a different compositor. Source photographs, objects, metadata, prompts, and scoring were unchanged; validation agreement was 88.2\% ($\kappa=0.728$). ECC labels follow from archive metadata and the post-1930 object criterion under human validation; however, this swap covers only compositing, so effects of model-assisted filtering, classification, and object selection may remain.

\noindent\textbf{Country-Balanced Robustness.}
Using stored predictions, we formed subsets with 40 pairs per country for Gemini composites ($N=320$) and 25 for GPT composites ($N=200$), the latter limited by accepted CN/JP pairs. Gemini variants retain the highest cECC on both (Table~\ref{tab:balanced_cecc}); this checks aggregate robustness, not country-specific judgments.

\noindent\textbf{Evidence Elicitation.}
\begin{table}[t]
\begin{minipage}[t]{0.585\linewidth}
\input{tables/table_07}
\end{minipage}
\hfill
\begin{minipage}[t]{0.39\linewidth}
\input{tables/table_09}

\end{minipage}
\end{table}
To examine the use of object evidence, we selected 84 hard pairs where all six selected models passed both IOP questions and original-image ECC, but at least four failed edited-image ECC. On each edited image, models returned a yes/no ECC judgment, an \texttt{Evidence} field, and a one-sentence \texttt{Reason}; Evidence ECC scores only the yes/no judgment. We checked object mentions by name, synonym, or clear visual description using keyword/synonym matching and manual verification (Table~\ref{tab:evidence_ecc}; Appendix~\ref{sec:appendix_evidence}).

The object appears in 165/175 correct responses (94.3\%) but only 28/329 incorrect responses (8.5\%). Llama-3.2-11B mentions it in 20/83 incorrect responses, treating it as compatible or answering \textit{Yes} after noting inconsistency. Correct judgments are strongly associated with using object evidence, although identifying it alone does not ensure a correct decision. Because all six models passed IOP on these pairs, these errors suggest that the perceived object is often not used in the whole-scene judgment.

\begin{table}[t]
\begin{minipage}[t]{0.57\linewidth}
\input{tables/table_08}
\end{minipage}
\hfill
\begin{minipage}[t]{0.40\linewidth}
\input{tables/table_10}
\end{minipage}
\end{table}
\looseness=-1
\noindent\textbf{Setting Identification.}
Table~\ref{tab:mcq_evidence_ecc} compares country/period identification through three-option multiple-choice questions (MCQs) on original images, using visual evidence alone, with Evidence ECC on edited images from the same 84 hard pairs. Country distractors are two other benchmark countries; period distractors are a nearby historical period and a modern period.

\looseness=-1
Country accuracy exceeds period accuracy by 15.5--40.5 points for every model. GPT-4o identifies both in 66.7\% of cases but reaches only 28.6\% Evidence ECC, suggesting distinct challenges in identifying a setting and applying it to consistency judgments.

\noindent\textbf{ECC Prompt Sensitivity.}
We sampled 20 pairs/country from 954 pairs where all six models passed both IOP questions, yielding 160 pairs selected independently of ECC. We tested the original prompt and two same-polarity paraphrases, fixing images, metadata, labels, and yes/no format (templates in Appendix~\ref{sec:appendix_paraphrase}). Under Paraphrase~1, five models change by at most 2.5 points and CogVLM2-19B by 10.6. The largest Paraphrase~2 changes are 13.8 for Llama-3.2-11B and 8.1 for CogVLM2-19B (Table~\ref{tab:prompt_cecc}). Stability varies by model; \textit{Yes} means consistent in all three prompts, so affirmative \mbox{bias is untested.}

\noindent\textbf{Object-Timing Knowledge Note.}
On the same 84 pairs, we added the object name and invention or widespread-adoption period to both ECC questions, using benchmark metadata cross-checked against Wikipedia. All other inputs and scoring were unchanged. Table~\ref{tab:knowledge_note} reports pairwise ECC: both images must be answered correctly (Appendix~\ref{sec:appendix_knowledge_note}).

GPT-4o improves from 10.7\% to 92.9\%, and both Gemini models exceed 91\%. InternVL3.5-14B improves more modestly, whereas Llama-3.2-11B remains unchanged and CogVLM2-19B declines. Explicit timing information thus resolves most hard pairs for GPT-4o and the Gemini models but not for Llama-3.2-11B or CogVLM2-19B, suggesting that models differ in using supplied object timing; this selected subset, however, does not separate missing knowledge from evidence selection.

%% file: tables/table_02.tex
\begin{table}[b]
\caption{Performance comparison across all eight countries on IOP and ECC (\%). \textbf{Bold} and \underline{underline} indicate the best and second-best model results in each column.}
\label{tab:main1}
\centering
\footnotesize
\setlength{\tabcolsep}{3pt}
\renewcommand{\arraystretch}{1.05}
\makebox[\linewidth][c]{%
\adjustbox{max width=1.0\linewidth}{%
\begin{tabular}{@{}l*{16}{c}@{}}
\toprule
\multirow{2}{*}{\textbf{Model}} & \multicolumn{2}{c}{\textbf{DE}} & \multicolumn{2}{c}{\textbf{US}} & \multicolumn{2}{c}{\textbf{NL}} & \multicolumn{2}{c}{\textbf{FR}} & \multicolumn{2}{c}{\textbf{RU}} & \multicolumn{2}{c}{\textbf{KR}} & \multicolumn{2}{c}{\textbf{CN}} & \multicolumn{2}{c}{\textbf{JP}} \\
\cmidrule(lr){2-3}\cmidrule(lr){4-5}\cmidrule(lr){6-7}\cmidrule(lr){8-9}\cmidrule(lr){10-11}\cmidrule(lr){12-13}\cmidrule(lr){14-15}\cmidrule(lr){16-17}
 & \textbf{IOP} & \textbf{ECC} & \textbf{IOP} & \textbf{ECC} & \textbf{IOP} & \textbf{ECC} & \textbf{IOP} & \textbf{ECC} & \textbf{IOP} & \textbf{ECC} & \textbf{IOP} & \textbf{ECC} & \textbf{IOP} & \textbf{ECC} & \textbf{IOP} & \textbf{ECC} \\
\midrule
Human & 100.0 & 89.2 & 99.2 & 80.8 & 95.0 & 67.5 & 100.0 & 91.7 & 99.2 & 89.2 & 100.0 & 94.2 & 100.0 & 98.3 & 100.0 & 95.0 \\
\midrule
\multicolumn{17}{c}{\textbf{Closed-source Models}} \\
\midrule
Claude-Sonnet-4.5 & 60.1 & 57.8 & 64.7 & 51.9 & 56.9 & 56.9 & 64.8 & 58.6 & 66.5 & 64.9 & 68.9 & 45.2 & 59.2 & 42.9 & 65.0 & 57.5 \\
GPT-4o & \second{87.3} & 68.5 & 82.0 & 56.8 & 79.3 & 62.1 & \second{88.6} & 70.5 & 90.8 & 73.5 & 91.5 & 67.2 & \best{89.8} & 55.1 & 87.5 & 77.5 \\
Gemini-2.5-Flash & 73.9 & \second{82.5} & 78.9 & \second{75.9} & 76.3 & \best{79.3} & 79.5 & \second{78.6} & 80.0 & \best{87.0} & 89.3 & \second{78.5} & 85.7 & \second{77.6} & 75.0 & \second{85.0} \\
Gemini-2.5-Pro & 85.5 & \best{82.8} & 80.5 & \best{77.4} & 78.4 & \second{78.9} & 84.3 & \best{79.5} & 85.4 & \second{84.9} & \best{95.5} & \best{84.2} & \second{87.8} & \best{87.8} & 82.5 & \best{87.5} \\
\midrule
\multicolumn{17}{c}{\textbf{Open-source Models}} \\
\midrule
InstructBLIP-7B & 72.3 & 0.0 & 75.9 & 0.0 & 70.7 & 0.0 & 74.3 & 0.0 & 77.3 & 0.0 & 76.8 & 0.0 & 61.2 & 0.0 & 57.5 & 0.0 \\
InstructBLIP-13B & 65.5 & 0.0 & 68.0 & 0.0 & 68.1 & 0.0 & 66.7 & 0.0 & 71.9 & 0.0 & 78.0 & 0.0 & 57.1 & 0.0 & 65.0 & 0.0 \\
mPLUG-Owl2-7B & 58.7 & 0.9 & 69.5 & 3.4 & 60.3 & 1.3 & 61.4 & 1.0 & 74.6 & 16.2 & 79.1 & 1.7 & 77.6 & 4.1 & 87.5 & 2.5 \\
LLaVA-1.5-13B & 66.4 & 4.5 & 76.7 & 8.3 & 74.1 & 1.3 & 73.8 & 4.8 & 85.9 & 15.7 & 81.9 & 5.6 & 73.5 & 12.2 & 75.0 & 7.5 \\
Gemma3-4B & 78.7 & 4.3 & 81.6 & 14.3 & 74.6 & 3.4 & 81.9 & 7.1 & 88.6 & 14.1 & 84.2 & 10.2 & \best{89.8} & 0.0 & 80.0 & 5.0 \\
LLaVA-1.5-7B & 62.8 & 4.5 & 74.4 & 16.2 & 70.3 & 3.4 & 69.0 & 5.2 & 77.3 & 21.1 & 74.6 & 9.6 & 67.3 & 20.4 & 80.0 & 0.0 \\
Qwen2.5-VL-3B & 75.3 & 9.3 & 75.9 & 29.7 & 75.0 & 3.9 & 80.5 & 13.3 & 79.5 & 23.8 & 90.4 & 17.5 & 75.5 & 14.3 & 67.5 & 5.0 \\
Qwen2.5-VL-7B & 74.1 & 24.0 & 77.1 & 38.0 & 70.3 & 31.0 & 75.2 & 31.4 & 82.2 & 34.6 & 81.4 & 15.8 & 79.6 & 12.2 & 72.5 & 12.5 \\
InternVL3.5-8B & 83.9 & 30.4 & \second{82.3} & 47.7 & \second{81.9} & 32.3 & 84.8 & 43.3 & 91.4 & 52.4 & 91.5 & 26.0 & 85.7 & 16.3 & 82.5 & 27.5 \\
Gemma3-27B & 83.2 & 38.3 & 80.1 & 43.6 & 78.9 & 45.7 & 83.8 & 45.2 & \second{93.0} & 57.8 & 84.2 & 35.0 & \best{89.8} & 28.6 & 87.5 & 30.0 \\
InternVL3.5-14B & 80.7 & 39.7 & 77.1 & 50.0 & 81.5 & 38.8 & 86.7 & 57.6 & 91.9 & 53.0 & 88.1 & 58.2 & 69.4 & 32.7 & 80.0 & 32.5 \\
Llama-3.2-11B & \best{90.7} & 17.0 & \best{93.2} & 25.2 & \best{92.2} & 10.3 & \best{95.7} & 19.5 & \best{94.6} & 31.9 & \second{92.1} & 23.7 & \second{87.8} & 16.3 & \best{92.5} & 17.5 \\
CogVLM2-19B & 76.9 & 48.1 & 71.8 & 43.2 & 72.0 & 47.8 & 77.1 & 53.3 & 84.3 & 58.4 & 85.3 & 62.7 & \best{89.8} & 36.7 & \best{92.5} & 45.0 \\
Gemma3-12B & 79.8 & 33.8 & 77.1 & 36.8 & 77.2 & 39.7 & 78.6 & 45.2 & 85.9 & 52.4 & 87.0 & 44.6 & 85.7 & 28.6 & \second{90.0} & 27.5 \\
\bottomrule
\end{tabular}
}
}
\end{table}

%% file: tables/table_03.tex
\begin{table}[t]
\caption{Country-wise and overall Conditional \textit{Era-Country Consistency} (\%).}
\label{tab:main2}
\centering
\footnotesize
\setlength{\tabcolsep}{5.75pt}
\renewcommand{\arraystretch}{1.05}
\makebox[\linewidth][c]{%
\adjustbox{max width=0.8\linewidth}{%
\begin{tabular}{lccccccccc}
\toprule
\textbf{Model} & \textbf{DE} & \textbf{US} & \textbf{NL} & \textbf{FR} & \textbf{RU} & \textbf{KR} & \textbf{CN} & \textbf{JP} & \textbf{Overall} \\
\midrule
Human & 89.2 & 81.4 & 71.6 & 91.7 & 89.9 & 94.2 & 98.3 & 95.0 & 88.9 \\
\midrule

\multicolumn{10}{c}{\textbf{Closed-source Models}} \\
\midrule
Claude-Sonnet-4.5 & 78.9 & 76.9 & 82.6 & 86.1 & 83.7 & 78.7 & 75.9 & 88.5 & 80.7 \\
GPT-4o & 76.0 & 66.4 & 72.6 & 76.6 & 79.3 & 72.4 & 61.4 & 83.3 & 73.8 \\
Gemini-2.5-Flash & \best{92.3} & \second{86.7} & \best{90.4} & \best{96.4} & \best{94.6} & \best{93.7} & \second{92.9} & \second{96.7} & \best{92.2} \\
Gemini-2.5-Pro & \second{91.5} & \textbf{86.9} & \second{89.6} & \second{93.8} & \second{93.0} & \second{89.9} & \best{95.3} & \best{97.0} & \second{91.1} \\
\midrule

\multicolumn{10}{c}{\textbf{Open-source Models}} \\
\midrule
InstructBLIP-7B & 0.0 & 0.0 & 0.0 & 0.0 & 0.0 & 0.0 & 0.0 & 0.0 & 0.0 \\
InstructBLIP-13B & 0.0 & 0.0 & 0.0 & 0.0 & 0.0 & 0.0 & 0.0 & 0.0 & 0.0 \\
mPLUG-Owl2-7B & 1.5 & 5.3 & 2.1 & 1.5 & 21.1 & 2.7 & 2.5 & 2.7 & 5.0 \\
LLaVA-1.5-13B & 5.5 & 10.6 & 1.6 & 5.5 & 18.4 & 6.9 & 12.5 & 9.4 & 8.0 \\
Gemma3-4B & 5.0 & 25.2 & 4.5 & 10.0 & 15.5 & 11.0 & 0.0 & 5.9 & 10.7 \\
LLaVA-1.5-7B & 6.2 & 21.5 & 4.7 & 6.0 & 27.2 & 12.6 & 25.7 & 0.0 & 12.4 \\
Qwen2.5-VL-3B & 12.6 & 42.6 & 5.1 & 15.4 & 28.4 & 19.6 & 18.4 & 3.6 & 19.5 \\
Qwen2.5-VL-7B & 28.4 & 55.6 & 41.1 & 36.5 & 40.1 & 17.8 & 17.5 & 16.7 & 35.3 \\
InternVL3.5-8B & 35.1 & 64.7 & 39.5 & 53.1 & 61.5 & 28.4 & 18.6 & 33.3 & 44.8 \\
Gemma3-27B & 42.3 & 50.4 & 51.8 & 51.6 & 60.5 & 37.8 & 32.6 & 31.6 & 47.2 \\
InternVL3.5-14B & 48.2 & 76.3 & 48.7 & 74.9 & 75.9 & 67.9 & 58.8 & 54.5 & 62.6 \\
Llama-3.2-11B & 18.7 & 33.1 & 12.0 & 19.6 & 35.6 & 24.7 & 15.6 & 17.9 & 22.8 \\
CogVLM2-19B & 55.2 & 59.7 & 60.7 & 62.6 & 78.8 & 82.8 & 54.5 & 48.6 & 63.7 \\
Gemma3-12B & 38.8 & 44.3 & 45.1 & 54.4 & 59.1 & 56.2 & 35.6 & 35.1 & 46.8 \\
\bottomrule
\end{tabular}%
}
}

\end{table}

%% file: tables/table_04.tex
\begin{table}[t]
\caption{Country-wise and overall conditional consistency accuracy (\%) on historically consistent controls, computed on edited images whose inserted era-appropriate object was perceived correctly.}
\label{tab:submain}
\centering
\footnotesize
\setlength{\tabcolsep}{4.2pt}
\renewcommand{\arraystretch}{1.05}
\makebox[\linewidth][c]{%
\adjustbox{max width=0.95\linewidth}{%
\begin{tabular}{l c c c c c c c c c}
\toprule
\textbf{Model} & \textbf{DE} & \textbf{US} & \textbf{NL} & \textbf{FR} & \textbf{RU} & \textbf{KR} & \textbf{CN} & \textbf{JP} & \textbf{Overall} \\
\midrule
Claude-Sonnet-4.5 & 89.6 & 87.0 & 94.7 & 81.4 & 77.2 & 65.3 & 82.1 & 76.9 & 84.3 \\
GPT-4o & 100.0 & 100.0 & 100.0 & 98.9 & 98.4 & 100.0 & 100.0 & 100.0 & 99.6 \\
Gemini-2.5-Flash & 94.5 & 96.6 & 98.8 & 87.6 & 90.3 & 88.0 & 96.4 & 92.9 & 93.5 \\
Gemini-2.5-Pro & 96.9 & 98.3 & 97.5 & 91.1 & 100.0 & 88.2 & 100.0 & 100.0 & 96.1 \\
\bottomrule
\end{tabular}%
}
}

\end{table}

%% file: figures/figure_03.tex
\begin{figure}[t]
    \centering
    \includegraphics[width=1.0\textwidth]{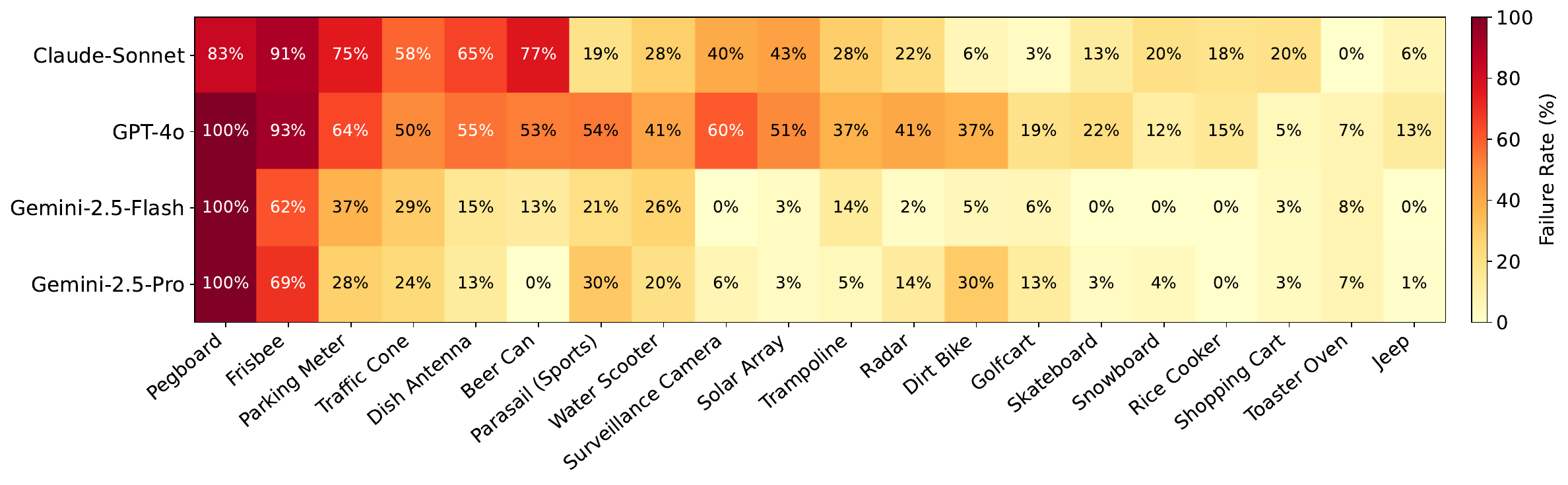}
    \setlength{\capinset}{0.05\linewidth}
    \caption{Object-level conditional ECC failure rates for the four closed-source models across the 20 most challenging inserted objects.}
    \label{fig:figure4}
\end{figure}

%% file: tables/table_05.tex
\caption{Conditional ECC (\%) on 432 accepted \texttt{gpt-image-1.5} composites for six of the 11 rerun models; denominators vary by model.}
\label{tab:main_gpt}
\centering
\footnotesize
\setlength{\tabcolsep}{10pt}
\renewcommand{\arraystretch}{1.12}
\makebox[\linewidth][c]{%
\adjustbox{max width=0.95\linewidth}{%
\begin{tabular}{lc}
\toprule
\textbf{Model} & \textbf{cECC (\%)} \\
\midrule
GPT-4o & 82.3 \\
Gemini-2.5-Flash & 94.6 \\
Gemini-2.5-Pro & 95.8 \\
InternVL3.5-14B & 72.8 \\
Llama-3.2-11B & 28.7 \\
CogVLM2-19B & 75.1 \\
\bottomrule
\end{tabular}
}
}

%% file: tables/table_06.tex
\caption{cECC (\%) on country-balanced subsets. $N$ is before IOP conditioning; denominators vary by model. Bold: column best.}
\label{tab:balanced_cecc}
\centering
\footnotesize
\setlength{\tabcolsep}{5pt}
\renewcommand{\arraystretch}{1.08}
\makebox[\linewidth][c]{%
\adjustbox{max width=1.0\linewidth}{%
\begin{tabular}{lcc}
\toprule
\textbf{Model} & \makecell{\textbf{Gemini subset}\\\textbf{($N=320$)}} & \makecell{\textbf{GPT subset}\\\textbf{($N=200$)}} \\
\midrule
GPT-4o & 74.4 & 82.5 \\
Gemini-2.5-Flash & \textbf{95.2} & 95.1 \\
Gemini-2.5-Pro & 94.7 & \textbf{97.3} \\
InternVL3.5-14B & 67.0 & 76.4 \\
Llama-3.2-11B & 22.1 & 29.6 \\
CogVLM2-19B & 63.6 & 77.6 \\
\bottomrule
\end{tabular}
}
}

%% file: tables/table_07.tex
\caption{Inserted-object mentions among correct/incorrect ECC responses on the 84 hard pairs. A dash indicates no correct response.}
\label{tab:evidence_ecc}
\centering
\footnotesize
\setlength{\tabcolsep}{4.2pt}
\renewcommand{\arraystretch}{1.08}
\makebox[\linewidth][c]{%
\adjustbox{max width=0.95\linewidth}{%
\begin{tabular}{lcc}
\toprule
\textbf{Model} & \textbf{Correct ECC} & \textbf{Incorrect ECC} \\
\midrule
GPT-4o & 23 / 24 & 0 / 60 \\
Gemini-2.5-Flash & 54 / 55 & 3 / 29 \\
Gemini-2.5-Pro & 67 / 69 & 0 / 15 \\
InternVL3.5-14B & 20 / 26 & 2 / 58 \\
Llama-3.2-11B & 1 / 1 & 20 / 83 \\
CogVLM2-19B & -- & 3 / 84 \\
\midrule
\textbf{Overall} & \textbf{165 / 175} & \textbf{28 / 329} \\
\bottomrule
\end{tabular}
}
}

%% file: tables/table_09.tex
\caption{Setting MCQ and Evidence ECC (\%) on 84 hard pairs. Both: correct country and period. Model names are abbreviated.}
\label{tab:mcq_evidence_ecc}
\centering
\footnotesize
\setlength{\tabcolsep}{3pt}
\renewcommand{\arraystretch}{1.08}
\makebox[\linewidth][c]{%
\adjustbox{max width=1.0\linewidth}{%
\begin{tabular}{lcccc}
\toprule
\multirow{2}{*}{\textbf{Model}} & \multicolumn{3}{c}{\textbf{Setting MCQ}} & \multirow{2}{*}{\makecell{\textbf{Evidence}\\\textbf{ECC}}} \\
\cmidrule(lr){2-4}
& \textbf{Ctry} & \textbf{Period} & \textbf{Both} & \\
\midrule
GPT-4o & 91.7 & 69.0 & 66.7 & 28.6 \\
Gemini-Flash & 97.6 & 71.4 & 69.0 & 65.5 \\
Gemini-Pro & 96.4 & 81.0 & 77.4 & 82.1 \\
InternVL-14B & 85.7 & 59.5 & 50.0 & 31.0 \\
Llama-11B & 67.9 & 48.8 & 38.1 & 1.2 \\
CogVLM2-19B & 72.6 & 32.1 & 20.2 & 0.0 \\
\bottomrule
\end{tabular}%
}
}

%% file: tables/table_08.tex
\caption{cECC (\%) under the original prompt and two same-polarity paraphrases on the same 160 IOP-correct pairs (20 per country).}
\label{tab:prompt_cecc}
\centering
\footnotesize
\setlength{\tabcolsep}{4.2pt}
\renewcommand{\arraystretch}{1.08}
\makebox[\linewidth][c]{%
\adjustbox{max width=1.0\linewidth}{%
\begin{tabular}{lccc}
\toprule
\textbf{Model} & \textbf{Original} & \textbf{Paraphrase 1} & \textbf{Paraphrase 2} \\
\midrule
GPT-4o & 85.0 & 86.9 & 80.6 \\
Gemini-2.5-Flash & 95.0 & 95.6 & 94.4 \\
Gemini-2.5-Pro & 96.3 & 96.3 & 96.9 \\
Llama-3.2-11B & 25.0 & 27.5 & 11.3 \\
InternVL3.5-14B & 67.5 & 65.0 & 65.6 \\
CogVLM2-19B & 73.8 & 63.1 & 65.6 \\
\bottomrule
\end{tabular}
}
}

%% file: tables/table_10.tex
\caption{Pairwise ECC (\%) on 84 hard pairs, without/with object-timing information. Differences are computed from counts.}
\label{tab:knowledge_note}
\centering
\footnotesize
\setlength{\tabcolsep}{5pt}
\renewcommand{\arraystretch}{1.08}
\makebox[\linewidth][c]{%
\adjustbox{max width=0.95\linewidth}{%
\begin{tabular}{lccc}
\toprule
\textbf{Model} & \textbf{Base} & \textbf{Note} & \textbf{$\Delta$ (pp)} \\
\midrule
GPT-4o & 10.7 & 92.9 & +82.1 \\
Gemini-Flash & 60.7 & 92.9 & +32.1 \\
Gemini-Pro & 72.6 & 91.7 & +19.0 \\
InternVL-14B & 7.1 & 19.0 & +11.9 \\
Llama-11B & 4.8 & 4.8 & 0.0 \\
CogVLM2-19B & 4.8 & 1.2 & $-$3.6 \\
\bottomrule
\end{tabular}%
}
}

%% file: sec/05_Conclusion.tex
\section{Conclusion}
\label{sec:conclusion}

We introduced AnachroBench, a benchmark for evaluating historical consistency through controlled modern-object insertions into historical images. Comprising 1,600 image pairs from eight countries, it evaluates \textit{Inserted Object Perception} (IOP), \textit{Era-Country Consistency} (ECC), and \textit{Conditional Era-Country Consistency} (cECC). Across 14 open-source and four closed-source LVLMs, we find that strong object perception does not necessarily lead to correct era-country consistency judgments, particularly for open-source models. Further analyses show substantial variation across inserted objects, while cross-compositor and country-balanced analyses indicate that the main performance differences persist. On the selected hard pairs, evidence and setting analyses suggest that models often struggle to apply perceived object evidence to the setting. Overall, AnachroBench provides a targeted diagnostic of false consistency judgments by separating inserted-object perception from historical-context judgments.

%% file: sec/06_Appendix.tex
\section{Limitations}
\label{sec:appendix_limitations}

AnachroBench is limited to eight countries and images from 1870--1929; it neither represents the full range of historical settings nor measures failure rates in spontaneous free-form generation. We also restrict the object pool to high-confidence temporal incompatibilities, objects invented or first widely adopted after 1930. We did not verify country-specific adoption timelines, so the benchmark omits subtler cases whose anachronicity depends on regional diffusion. Covering those would require expert historical annotation. The binary format supports consistent quantitative evaluation but does not capture how models arrive at their judgments. The 45 objects are unevenly distributed because some integrate into historical scenes more plausibly than others; future versions should broaden the pool to include categories with a wider range of visual attributes. Finally, model-assisted filtering and insertion may introduce systematic biases despite the human validation used to screen the resulting images.

\section{Prompts Used in Benchmark Construction}
\label{sec:appendix_prompt}

Figures~\ref{fig:prompt_filtering}--\ref{fig:prompt_editing} provide the prompts for image filtering, object classification, object--image matching, and object insertion.

\begin{figure}[!htbp]
\centering
\begin{tcolorbox}[colback=gray!8, colframe=gray!40, width=0.9\linewidth, top=1.1mm, bottom=1.1mm]
\small

\textbf{System:} You are a strict historical photo curator. Output only a single JSON object, no markdown.

\medskip
\noindent\rule{\linewidth}{0.4pt}
\medskip

\textbf{User:} Evaluate the image based on the geographic region label: \texttt{\{region\_label\}}.

\medskip
Must not contain any graphic violence, dead or injured persons, executions, armed children, child soldiers, or any other disturbing, exploitative content.

\medskip
Decide if ALL the following criteria are clearly met based on VISIBLE evidence:

\medskip
1) \textbf{Region Match:} The scene must match the labeled region/country (\texttt{\{region\_label\}}) through architecture, signage, landscape, or cultural cues.\\[5pt]
2) \textbf{Era Window:} The depicted era must fall between \texttt{\{year\_min\}} and \texttt{\{year\_max\}}.\\[5pt]
3) \textbf{Attire as Temporal Cues:} Visible clothing must clearly align with the \texttt{\{year\_min\}}--\texttt{\{year\_max\}} era.

\end{tcolorbox}
\setlength{\capinset}{0.09\linewidth}
\caption{Image filtering prompt used with \texttt{gemini-3-flash-preview}. Region label and year bounds are substituted per image.}
\label{fig:prompt_filtering}
\end{figure}
\begin{figure}[!htbp]
\centering
\begin{tcolorbox}[colback=gray!8, colframe=gray!40, width=0.9\linewidth, top=1.1mm, bottom=1.1mm]
\small

\textbf{User:} You are a strict historical-object classifier. For each object name, decide whether it was invented or first widely adopted AFTER 1930.

\end{tcolorbox}
\setlength{\capinset}{0.09\linewidth}
\caption{Anachronistic object classification prompt used with \texttt{gemini-3.1-pro-preview}. Object names from LVIS and Objects365 are passed as input.}
\label{fig:prompt_classification}
\end{figure}
\begin{figure}[!htbp]
\centering
\begin{tcolorbox}[colback=gray!8, colframe=gray!40, width=0.9\linewidth, top=1.1mm, bottom=1.1mm]
\small

\textbf{User:} [image]

\medskip
You are selecting ONE object for historical-photo editing. Choose the single candidate that would look the MOST natural in this scene (scale, context, era contrast, plausible placement).

\end{tcolorbox}
\setlength{\capinset}{0.09\linewidth}
\caption{Object--image matching prompt used with \texttt{gemini-3-flash-preview}. The candidate list and image are provided per query.}
\label{fig:prompt_selection}
\end{figure}
\begin{figure}[!htbp]
\centering
\begin{tcolorbox}[colback=gray!8, colframe=gray!40, width=0.9\linewidth, top=1.1mm, bottom=1.1mm]
\small

\textbf{User:} Seamlessly insert exactly one \{object\} at a physically plausible location, keeping it naturally integrated into the scene.

\medskip
Preserve the original composition, people, lighting, and historical style strictly unchanged.

\medskip
Ensure accurate perspective for realistic blending, using flat lighting to reduce unnatural 3D dimensionality.

\end{tcolorbox}
\setlength{\capinset}{0.09\linewidth}
\caption{Object insertion prompt used with \texttt{gemini-2.5-flash-image}. The object name is substituted per image.}
\label{fig:prompt_editing}
\end{figure}

\section{AnachroBench}
\label{sec:appendix_anachroBench}

\subsection{Inter-Annotator Agreement}
\label{sec:appendix_iaa}

Table~\ref{tab:iaa} reports agreement for the human validation stage. Country-level $\kappa$ ranges from 0.544 to 0.953, spanning moderate to almost perfect agreement on the Landis--Koch scale~\cite{landis1977measurement}; the lowest estimate is for China ($n=136$). Only candidates independently accepted by both annotators entered the benchmark.

\paragraph{Annotation guideline.}
Annotators first verified that the inserted object was incompatible with the stated era and country. They then checked its placement, scale and perspective, color and photographic style, and any visible compositing artifacts. Failure on any criterion resulted in rejection. Disagreements arose mainly on the visual criteria, most often the object's size and placement. Historical incompatibility was rarely at issue, since the post-1930 pool leaves it unambiguous. We report this qualitatively rather than as counts.

\input{tables/benchmark_iaa}

\subsection{Dataset Statistics}
\label{sec:appendix_statistics}

\begin{figure}[!htbp]
\centering
\includegraphics[width=0.65\linewidth]{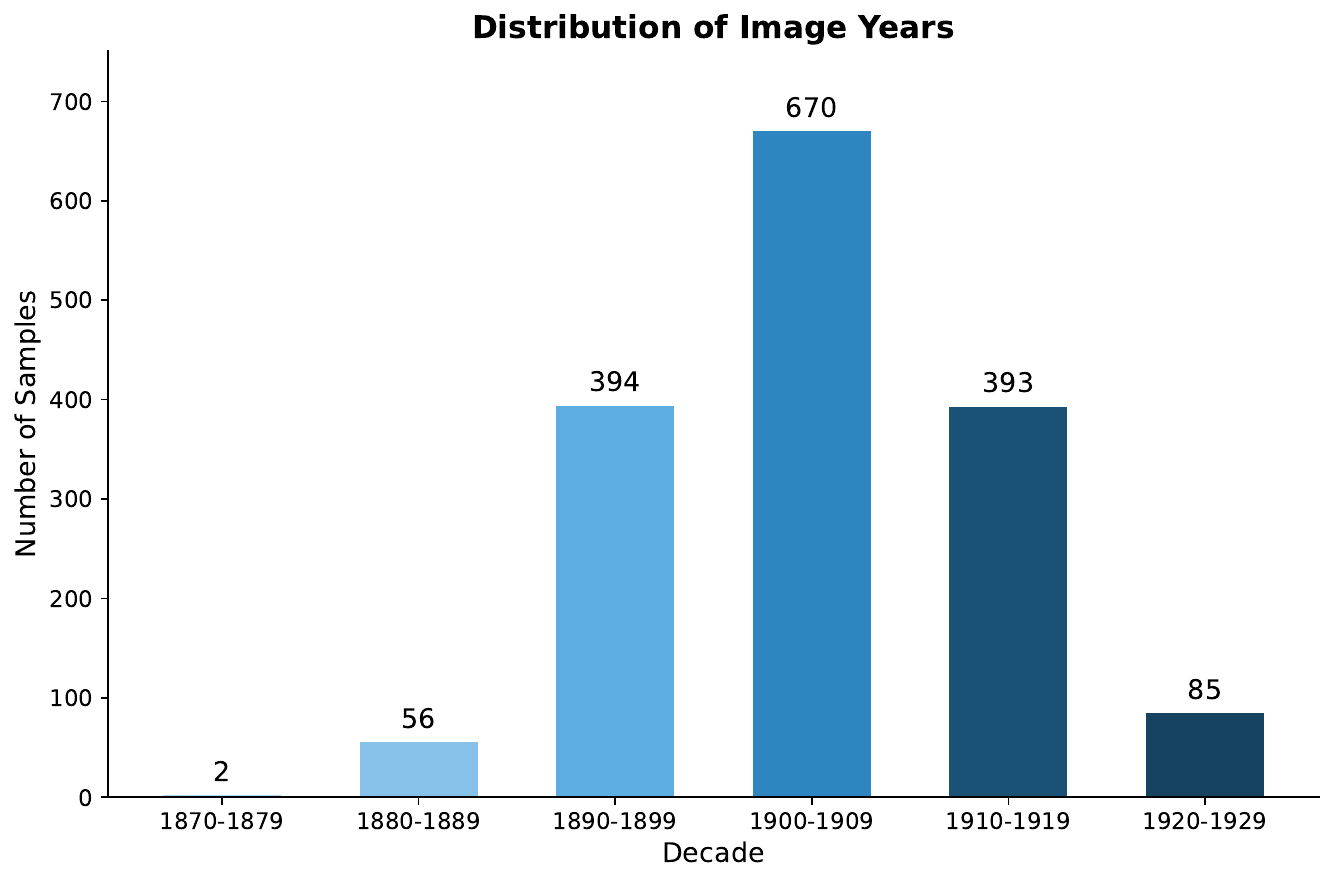}
\caption{Distribution of image years by decade across the benchmark.}
\label{fig:year}
\end{figure}

\input{tables/pre_filter_source}

\input{tables/benchmark_source}

\begin{figure}[!htbp]
\centering
\includegraphics[width=0.95\linewidth]{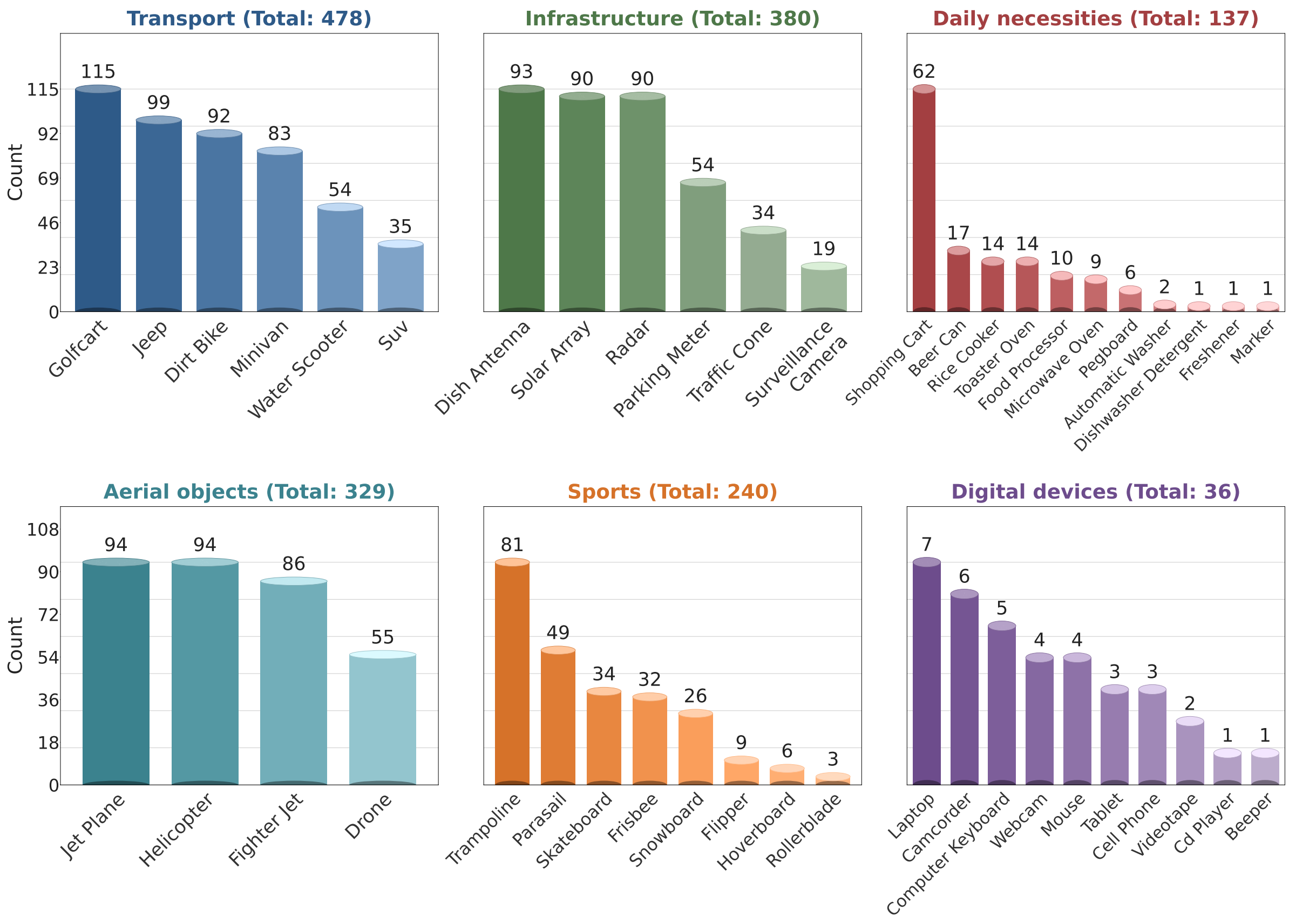}
\caption{Frequency of 45 anachronistic objects across the 1,600-pair benchmark, sorted in descending order.}
\label{fig:objects_categories}
\end{figure}

Tables~\ref{tab:pre_filter_source} and~\ref{tab:source} summarize the source- and country-wise image distributions before filtering and after filtering and human validation, respectively. Figure~\ref{fig:year} shows the temporal distribution: 670 images (41.9\%) date to 1900--1909, 452 (28.3\%) to before 1900, 393 (24.6\%) to 1910--1919, and 85 (5.3\%) to the 1920s. Figure~\ref{fig:objects_categories} gives the frequency of all 45 objects. Golf cart occurs 115 times, jeep 99 times, and jet plane and helicopter 94 times each; CD player, beeper, marker, freshener, and dishwasher detergent occur once each.

\FloatBarrier

\section{Historically Consistent Image Dataset}
\label{sec:appendix_consistent}

\subsection{Inter-Annotator Agreement}
\label{sec:appendix_iaa_sub}

\looseness=-1
Table~\ref{tab:iaa_sub} reports agreement for the historically consistent control set. Country-level $\kappa$ ranges from 0.576 to 0.773, spanning moderate to substantial agreement on the Landis--Koch scale~\cite{landis1977measurement}. Only images accepted by both annotators were retained.

\input{tables/control_iaa}

\subsection{Dataset Statistics}
\label{sec:appendix_sub}

\begin{figure}[!htbp]
\centering
\begin{minipage}[b]{0.47\linewidth}
\centering
\includegraphics[width=0.98\linewidth]{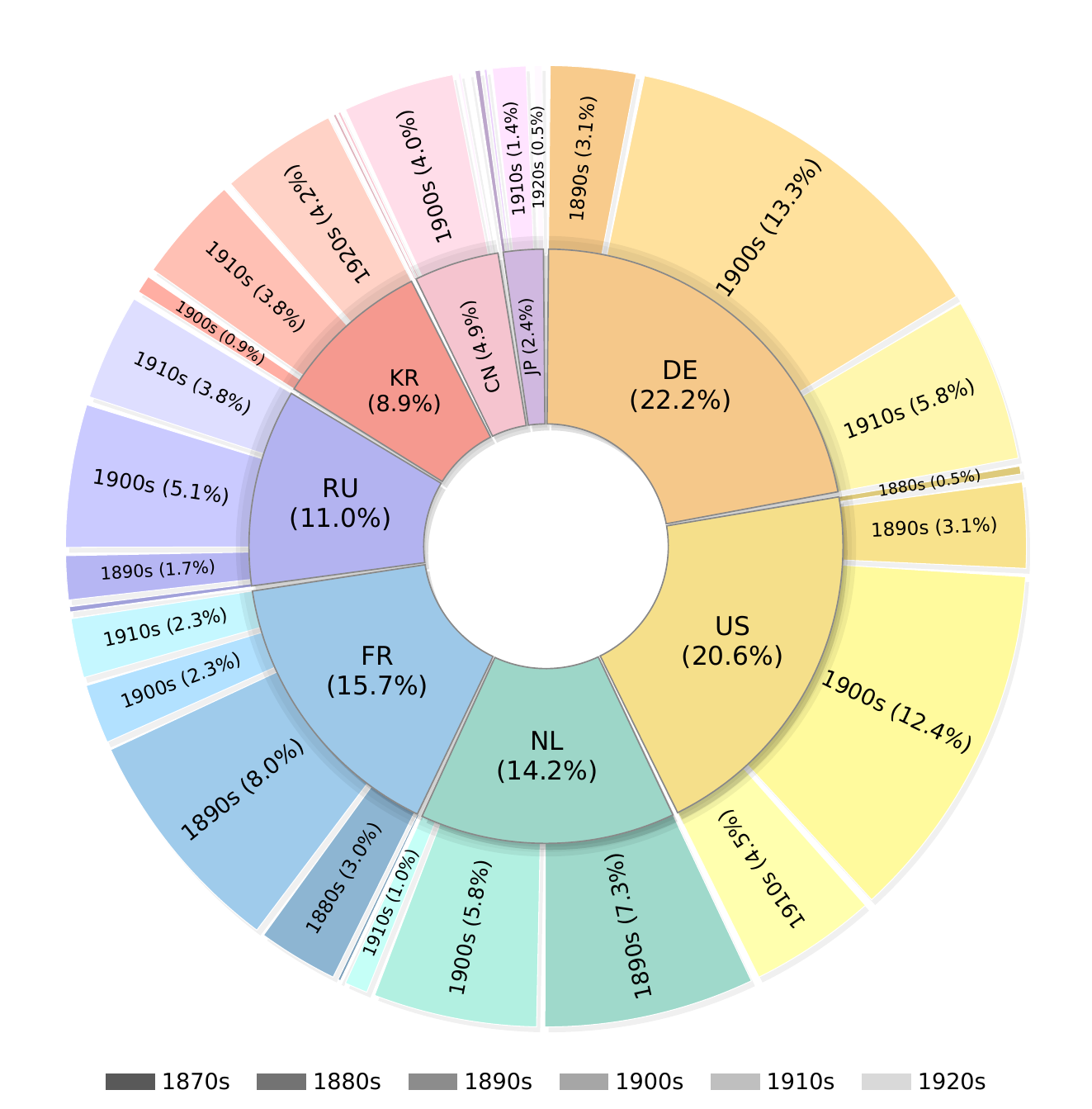}
\caption{Distribution of 572 historically consistent images by country (inner ring) and decade (outer ring).}
\label{fig:sub_sunburst}
\end{minipage}\hfill
\begin{minipage}[b]{0.51\linewidth}
\centering
\includegraphics[width=0.88\linewidth]{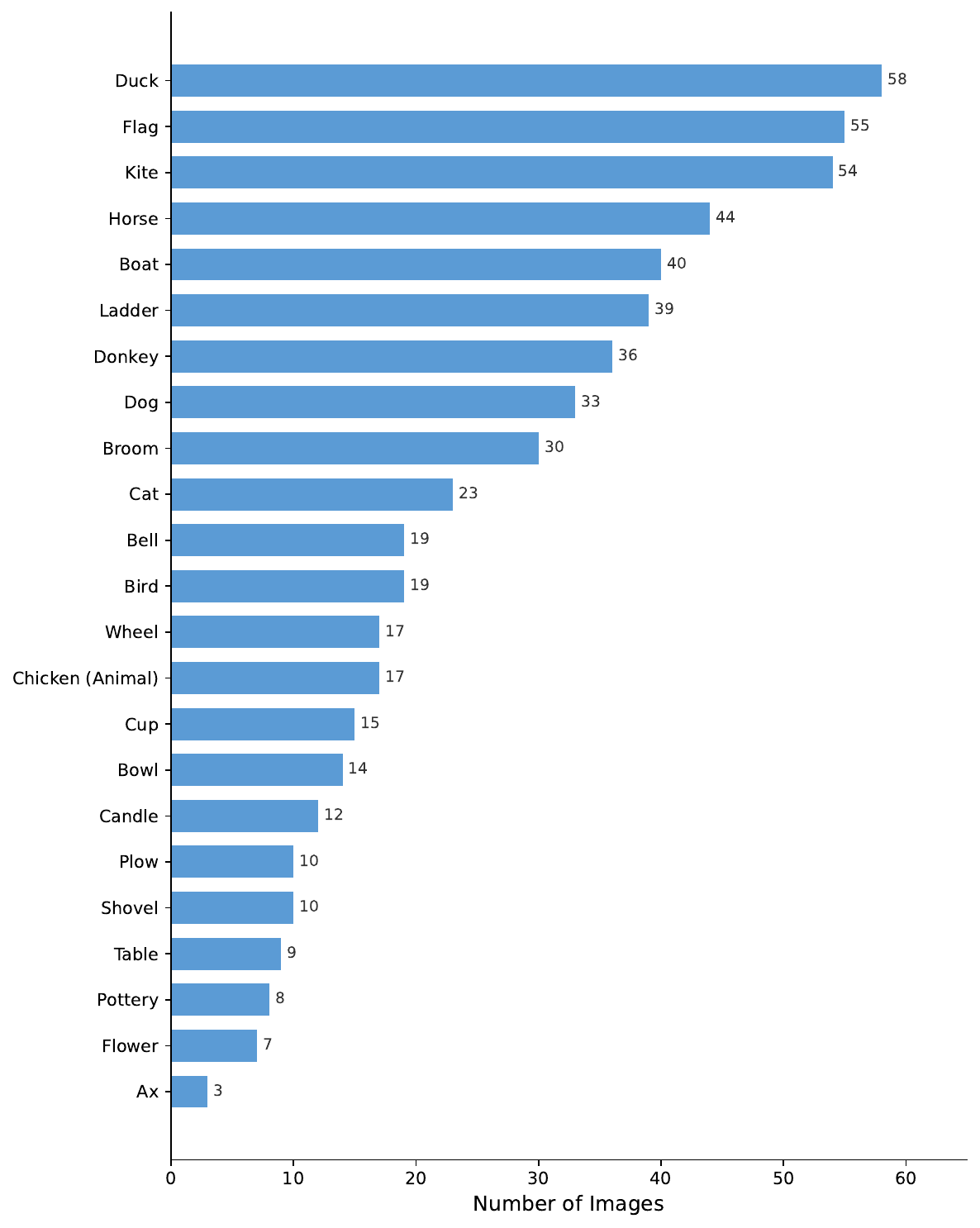}
\caption{Frequencies of 23 era-appropriate objects in 572 historically consistent images, sorted in descending order.}
\label{fig:sub_object_freq}
\end{minipage}
\end{figure}

Figure~\ref{fig:sub_sunburst} breaks down the 572 control images by country and decade; Figure~\ref{fig:sub_object_freq} shows the frequencies of the \mbox{23 era-appropriate objects.}

\FloatBarrier

\section{Additional Robustness and Diagnostic Analyses}
\label{sec:appendix_additional}

The analyses below use different subsets and protocols. Their results are therefore interpreted within each stated setting rather than as direct estimates on the full benchmark.

\subsection{Original--Anachronistic Performance by Country}
\label{sec:appendix_radar}

\begin{figure}[!htbp]
\centering
\includegraphics[width=0.95\linewidth]{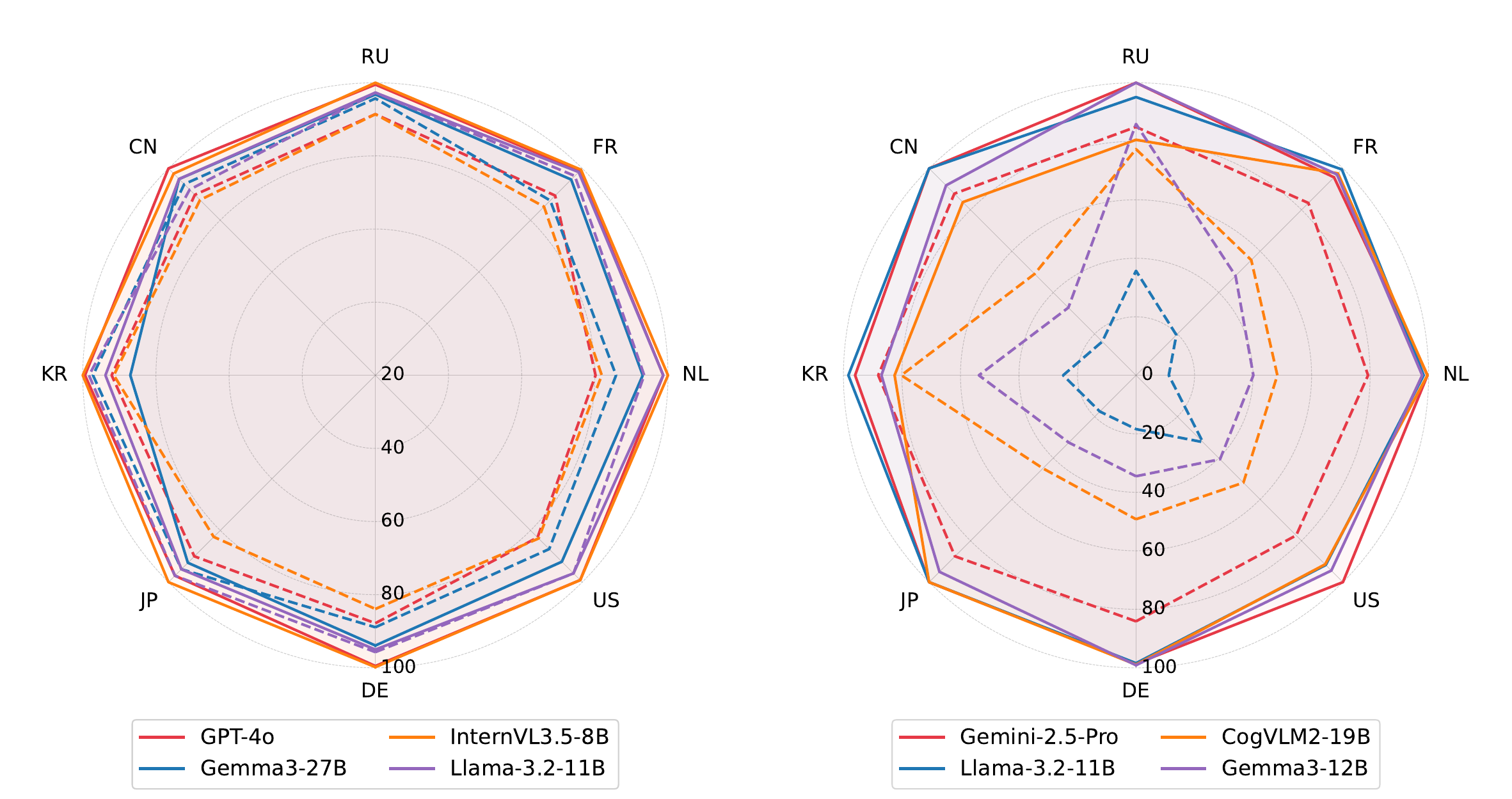}
\caption{Radar charts illustrating the performance of selected closed- and open-source models across eight countries. The left panel presents IOP, while the right panel presents ECC. Solid lines denote performance on original images, whereas dashed lines denote performance on the corresponding anachronistic images.}
\label{fig:radar_country}
\end{figure}

Figure~\ref{fig:radar_country} compares performance on original and anachronistic images across countries. IOP changes relatively little between the two image types, whereas ECC decreases substantially on anachronistic images, particularly for several open-source models. This gap indicates that the primary difficulty emerges when models must relate perceived objects to the specified historical setting.

\subsection{Cross-Compositor Robustness}
\label{sec:appendix_gpt_compositing}

To examine whether the results persist with a different compositor, we regenerated 689 pairs with \texttt{gpt-image-1.5}. The source photographs, inserted objects, metadata, prompts, and scoring protocol were unchanged, and both annotators accepted 432 pairs. Agreement was 88.2\% with Cohen's $\kappa = 0.728$, and country-level $\kappa$ ranged from 0.629 to 0.835. We reran 11 models on this set; Table~\ref{tab:gpt_compositing} reports their country-wise and overall conditional ECC (cECC).

\input{tables/gpt_compositing_full}

The two Gemini models retain the highest overall cECC among the rerun models. This result is descriptive rather than an estimate of a compositor effect: image filtering, object selection, and object classification remain Gemini-assisted, and the Gemini-composited baseline was not recomputed on the same 432 items. The country cells are also much smaller than those in the full benchmark and are not used for cross-country ranking.

\subsection{Setting Identification and Evidence Elicitation}
\label{sec:appendix_evidence}

For a closer view of false consistency judgments, we selected 84 pairs on which all six models passed IOP for both images and answered ECC correctly for the original image, but at least four failed ECC for the edited image. This shared hard subset is used for failure analysis, not to estimate full-benchmark performance.

Each model first answered separate three-option questions about the country and period of the original image using visual evidence alone. The country choices consist of the correct answer and two other benchmark countries; the period choices include the correct period, a nearby historical period, and a modern decoy. In a separate call on the edited image, the model returned a yes/no ECC judgment, an \texttt{Evidence} field, and a one-sentence \texttt{Reason}. Evidence ECC scores only the yes/no judgment, not the accompanying text. The two diagnostics are reported in Section~\ref{sec:analysis} and Table~\ref{tab:mcq_evidence_ecc}.

The setting questions and Evidence ECC use different images and response formats, so the columns of Table~\ref{tab:mcq_evidence_ecc} describe complementary behavior on the selected cases rather than a common capability scale.

The inserted-object mention audit on these Evidence ECC responses is reported in Section~\ref{sec:analysis} and Table~\ref{tab:evidence_ecc}.

\subsection{Object-Timing Knowledge Note}
\label{sec:appendix_knowledge_note}

On the same 84 pairs, we repeated the standard pairwise ECC evaluation after adding an item-specific note to both the original- and edited-image questions. The note names the inserted object and states when it was invented or first widely adopted, using benchmark metadata cross-checked against Wikipedia. Images, questions, answer polarity, and scoring remain unchanged. Table~\ref{tab:knowledge_note} compares the original prompt (Base) with this controlled information-provision condition (Note); it does not evaluate retrieval.

With the note, pair accuracy rises from 10.7\% to 92.9\% for GPT-4o and reaches 91.7--92.9\% for the two Gemini models. The smaller gain for InternVL3.5-14B, unchanged Llama-3.2-11B result, and decrease for CogVLM2-19B show that the response is not uniform. Because the note supplies both the object identity and its timing, the gains do not separate missing object-timing knowledge from failure to use relevant knowledge in the judgment. These pairwise ECC scores are distinct from the edited-image Evidence ECC.

\subsection{ECC Prompt Sensitivity}
\label{sec:appendix_paraphrase}

\input{tables/prompt_templates}

For the prompt-sensitivity analysis in Section~\ref{sec:analysis}, we evaluate the original ECC prompt and two same-polarity paraphrases on a fixed subset of 160 pairs, 20 per country (results in Table~\ref{tab:prompt_cecc}). Table~\ref{tab:prompt_templates} provides the exact prompt templates used in this analysis.

\subsection{Country and Object Breakdowns}
\label{sec:appendix_breakdowns}

Table~\ref{tab:aggregate_breakdowns} averages results over all 18 models. Country values use the full 1,600-pair benchmark.

\input{tables/aggregate_breakdowns}

IOP exceeds ECC in every country, although China and Japan each contain fewer than 50 pairs. Among the retained objects, cECC ranges from 62.9\% for fighter jets to 10.3\% for frisbees. These descriptive aggregates neither identify the source of the errors nor support fine-grained country rankings.

In Tables~\ref{tab:main1}--\ref{tab:main2}, the human references come from 40 pairs per country rather than the full 1,600 pairs. Because the evaluated sets and cECC denominators differ, they serve as descriptive references rather than a ceiling.

\section{Qualitative Examples}
\label{sec:appendix_examples}

\begin{figure}[!ht]
\centering
\setlength{\tabcolsep}{2pt}
\renewcommand{\arraystretch}{1.0}
\small

\begin{tabular}{@{}rccc@{}}
& \textbf{Original}
& \textbf{Anachronistic}
& \textbf{Historically Consistent} \\[6pt]

\rotatebox{90}{\shortstack{DE\\(1903)}} &
\includegraphics[width=0.305\textwidth]{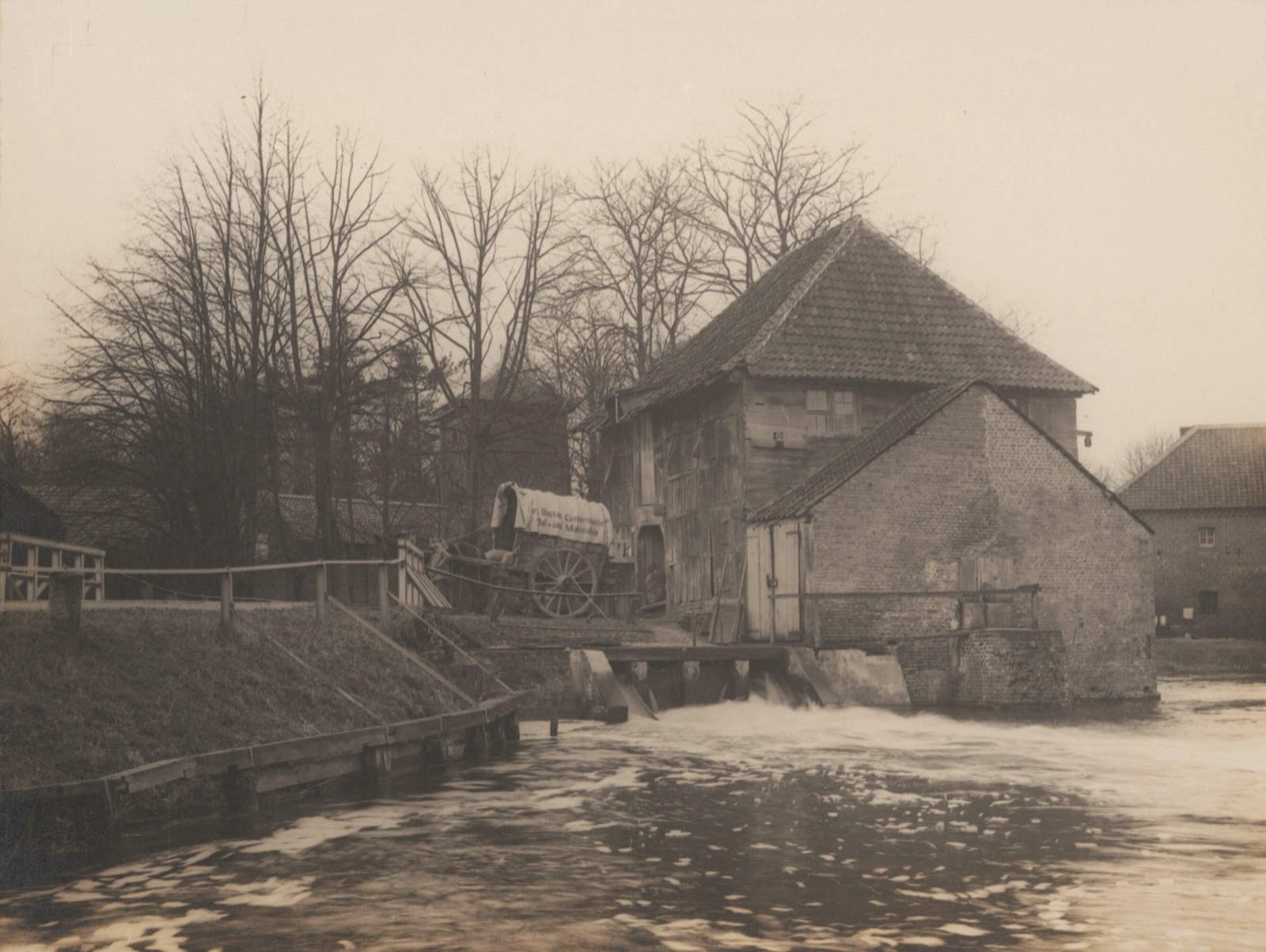} &
\includegraphics[width=0.305\textwidth]{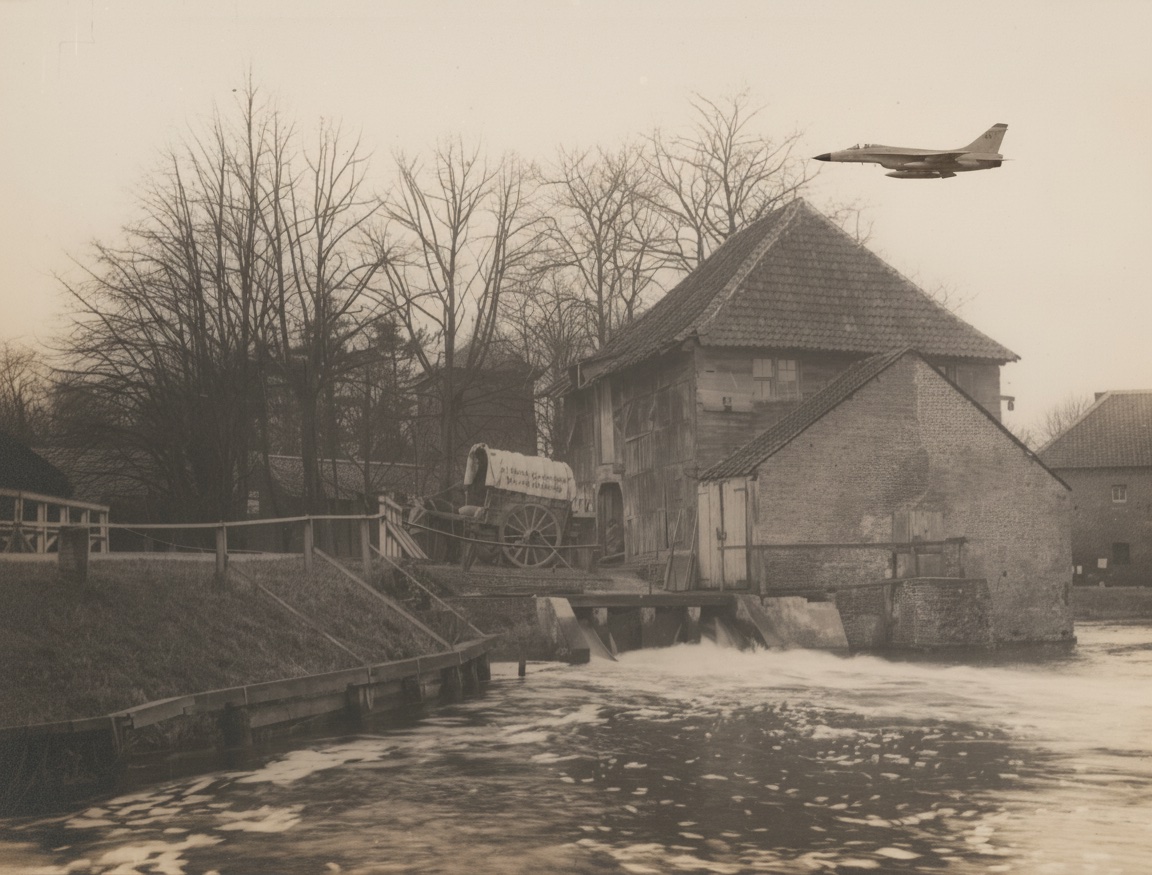} &
\includegraphics[width=0.305\textwidth]{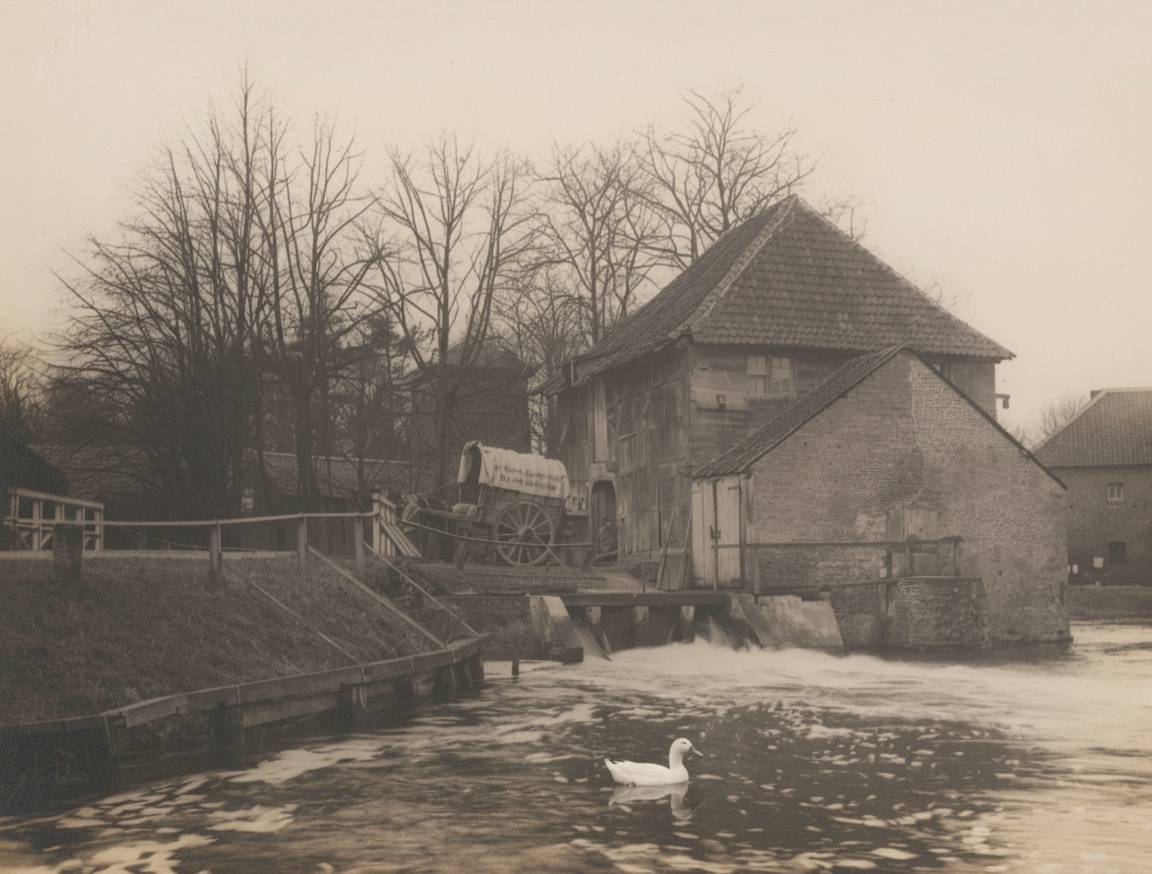} \\[-1pt]
& & \small\textit{Fighter Jet} & \small\textit{Duck} \\[10pt]

\rotatebox{90}{\shortstack{US\\(1901)}} &
\includegraphics[width=0.305\textwidth]{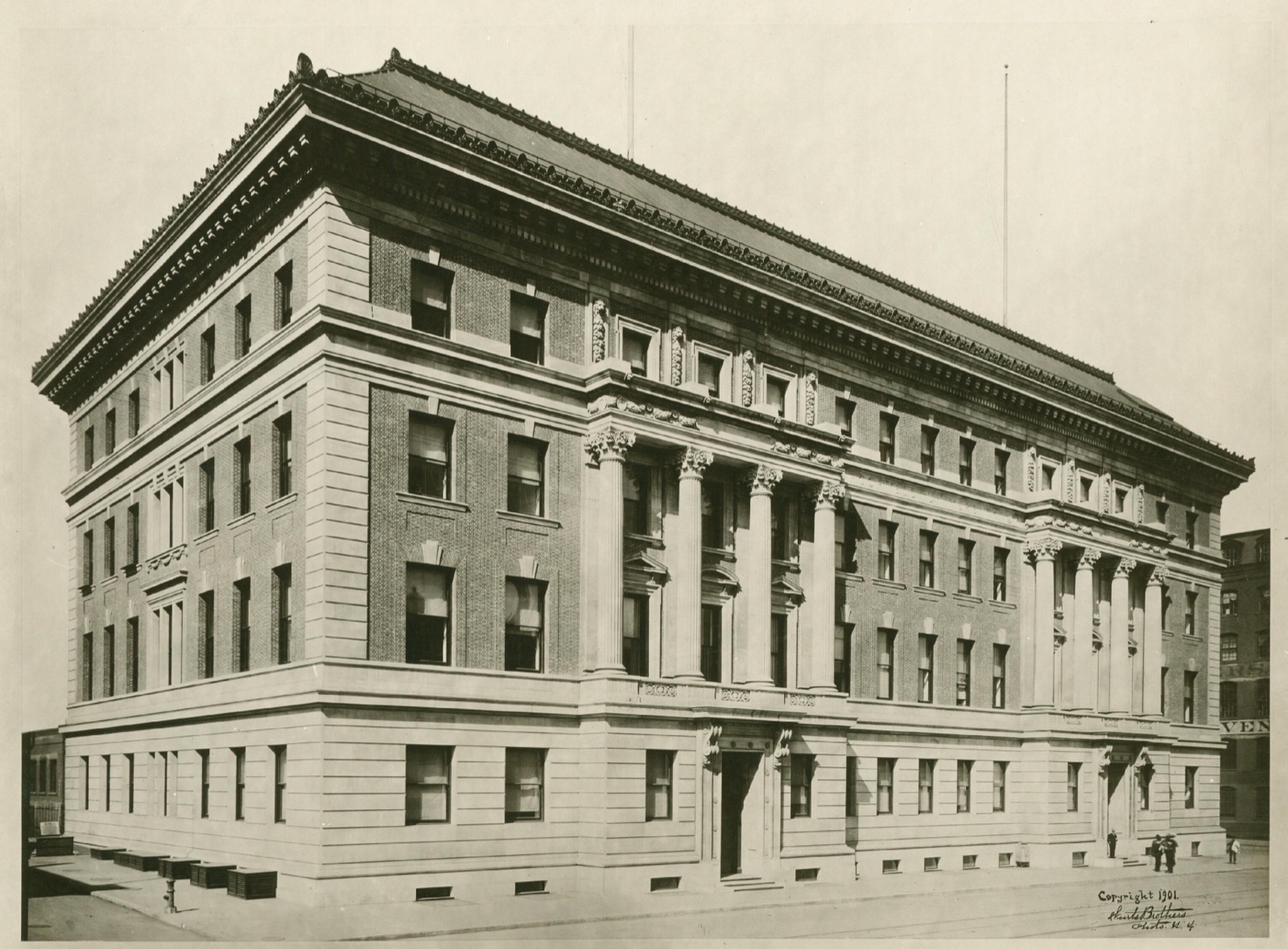} &
\includegraphics[width=0.305\textwidth]{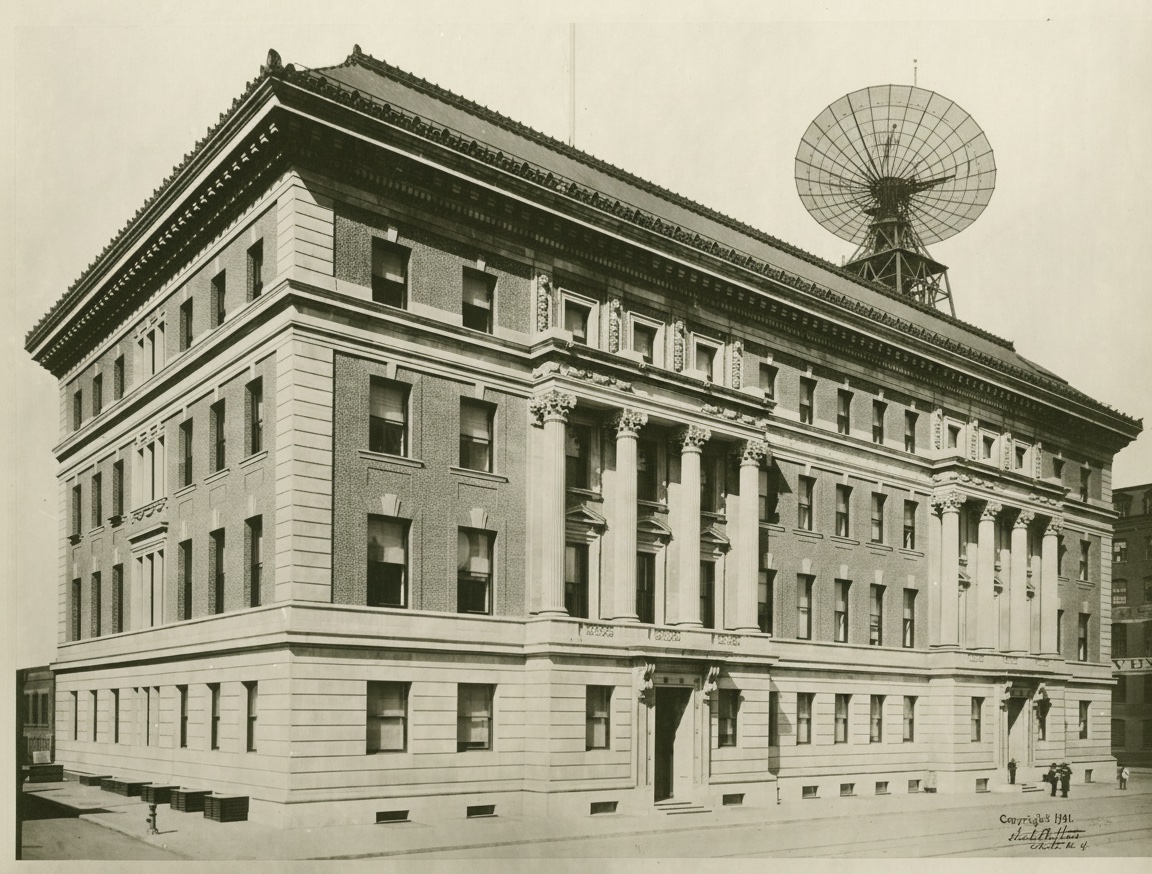} &
\includegraphics[width=0.305\textwidth]{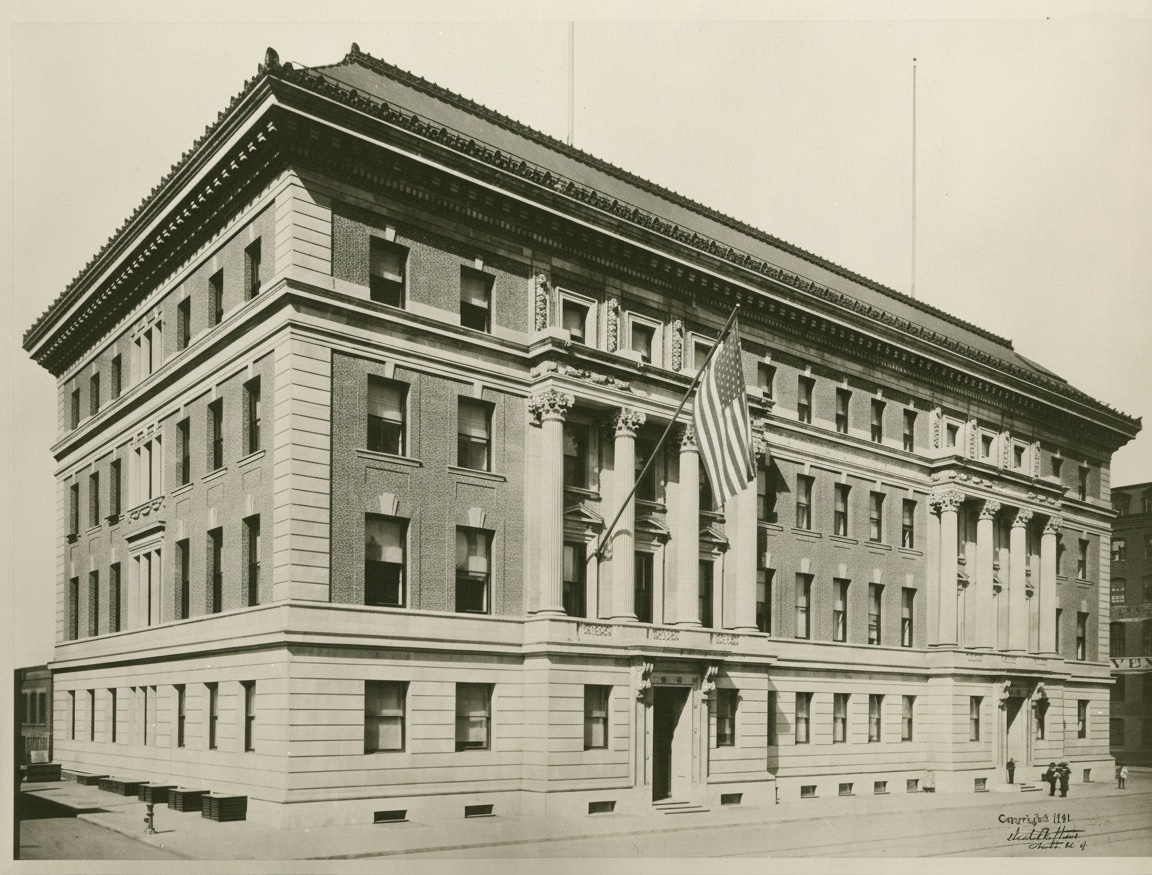} \\[-1pt]
& & \small\textit{Radar} & \small\textit{Flag} \\[10pt]

\rotatebox{90}{\shortstack{NL\\(1914)}} &
\includegraphics[width=0.305\textwidth]{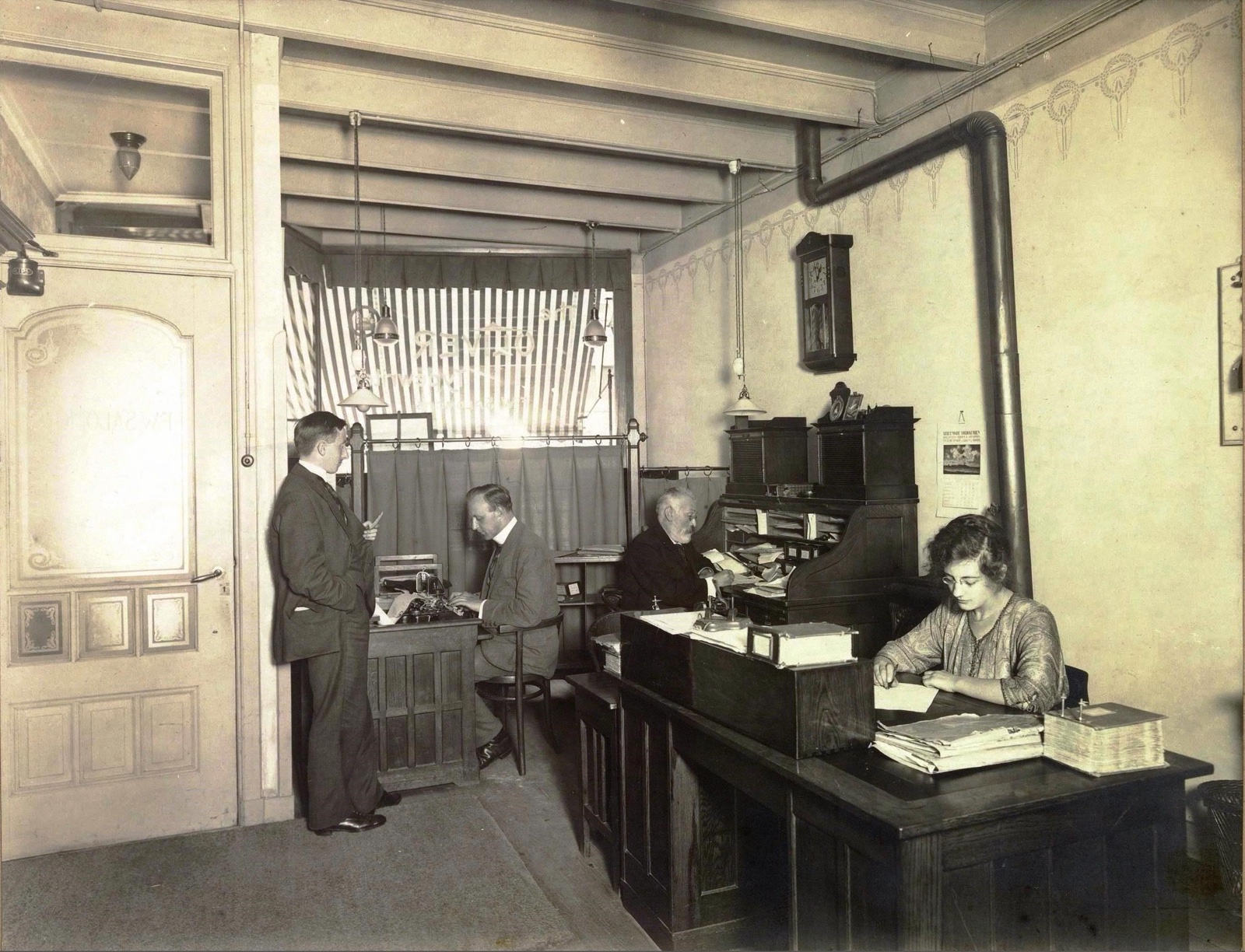} &
\includegraphics[width=0.305\textwidth]{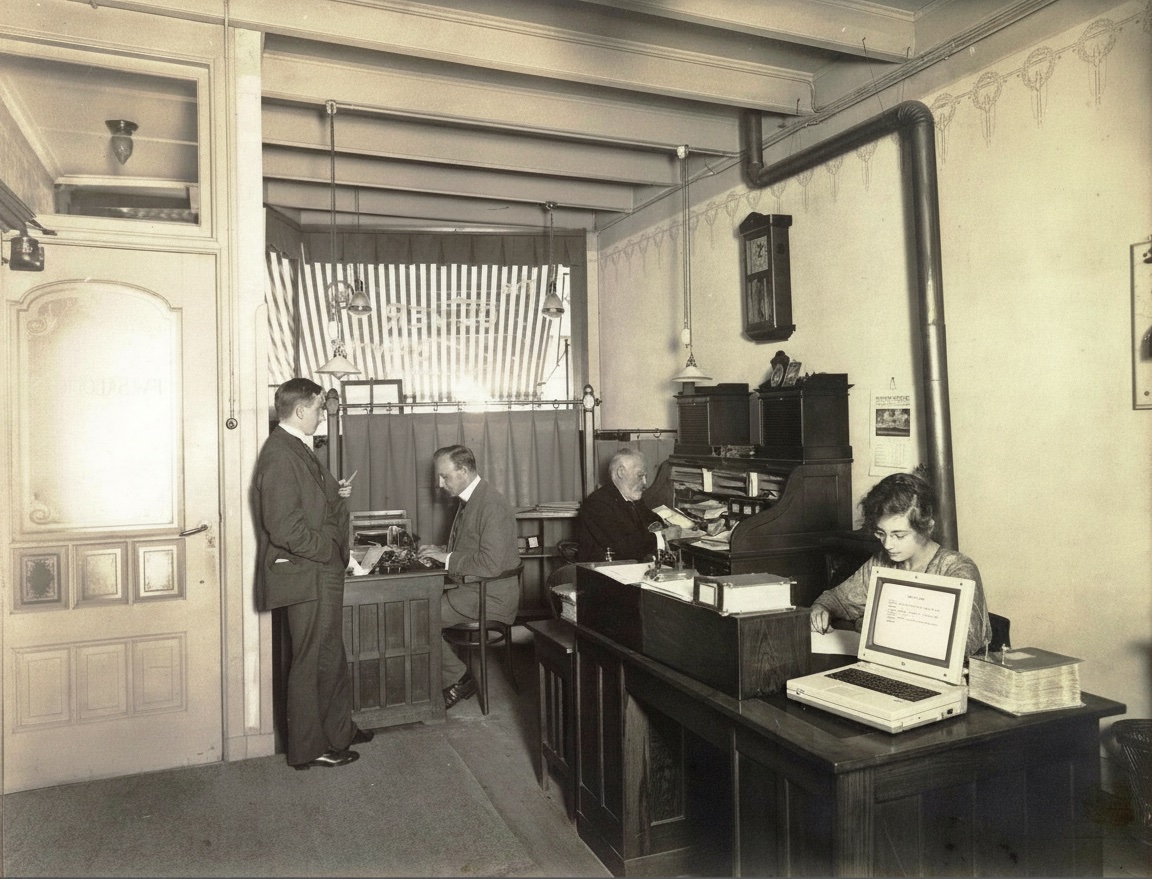} &
\includegraphics[width=0.305\textwidth]{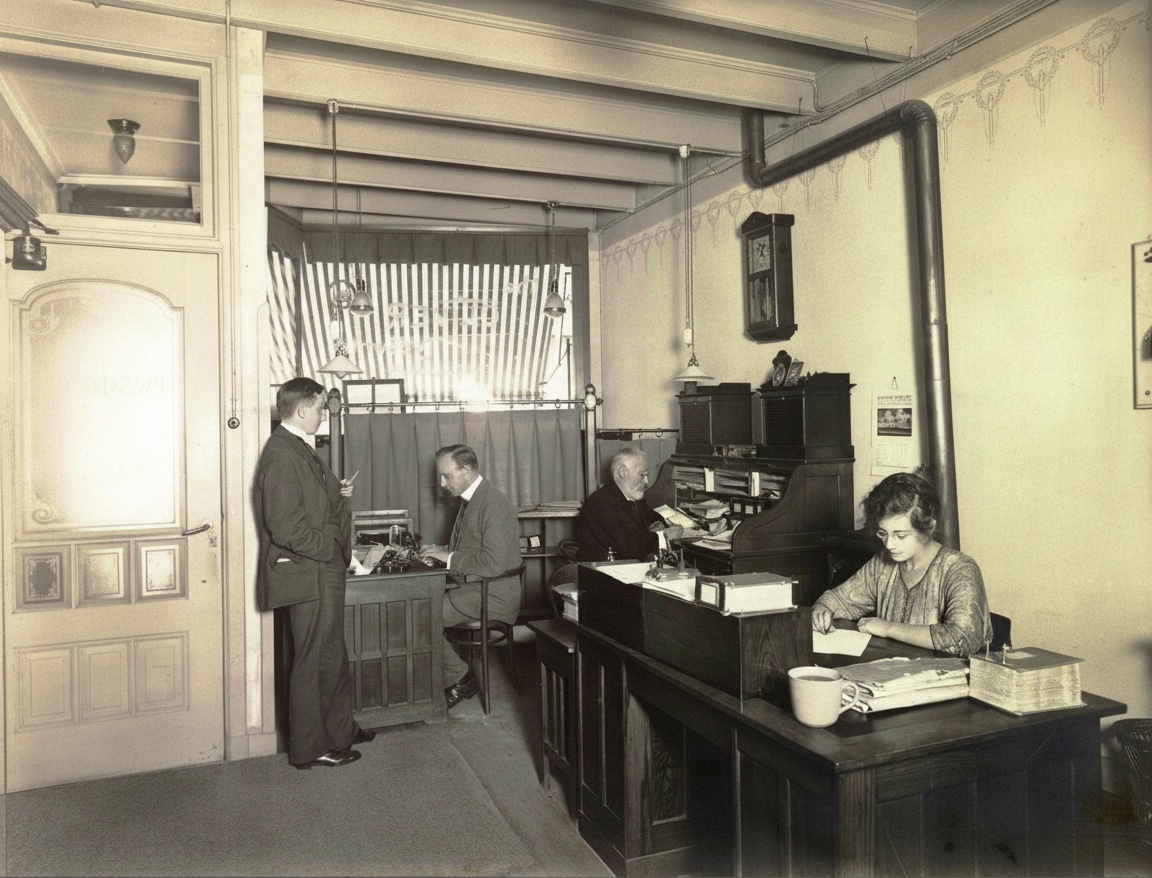} \\[-1pt]
& & \small\textit{Laptop} & \small\textit{Cup} \\[10pt]

\rotatebox{90}{\shortstack{FR\\(1897)}} &
\includegraphics[width=0.305\textwidth]{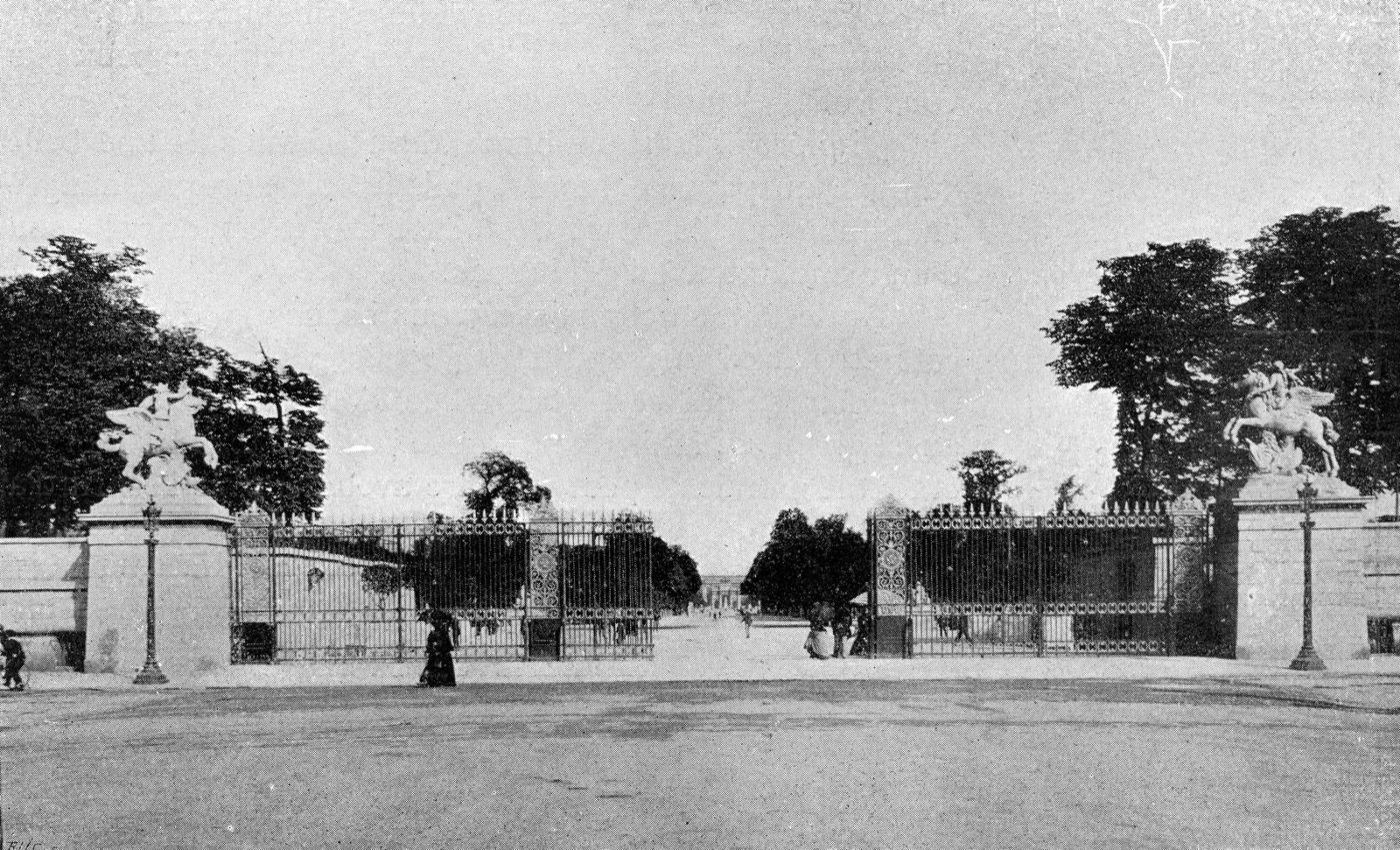} &
\includegraphics[width=0.305\textwidth]{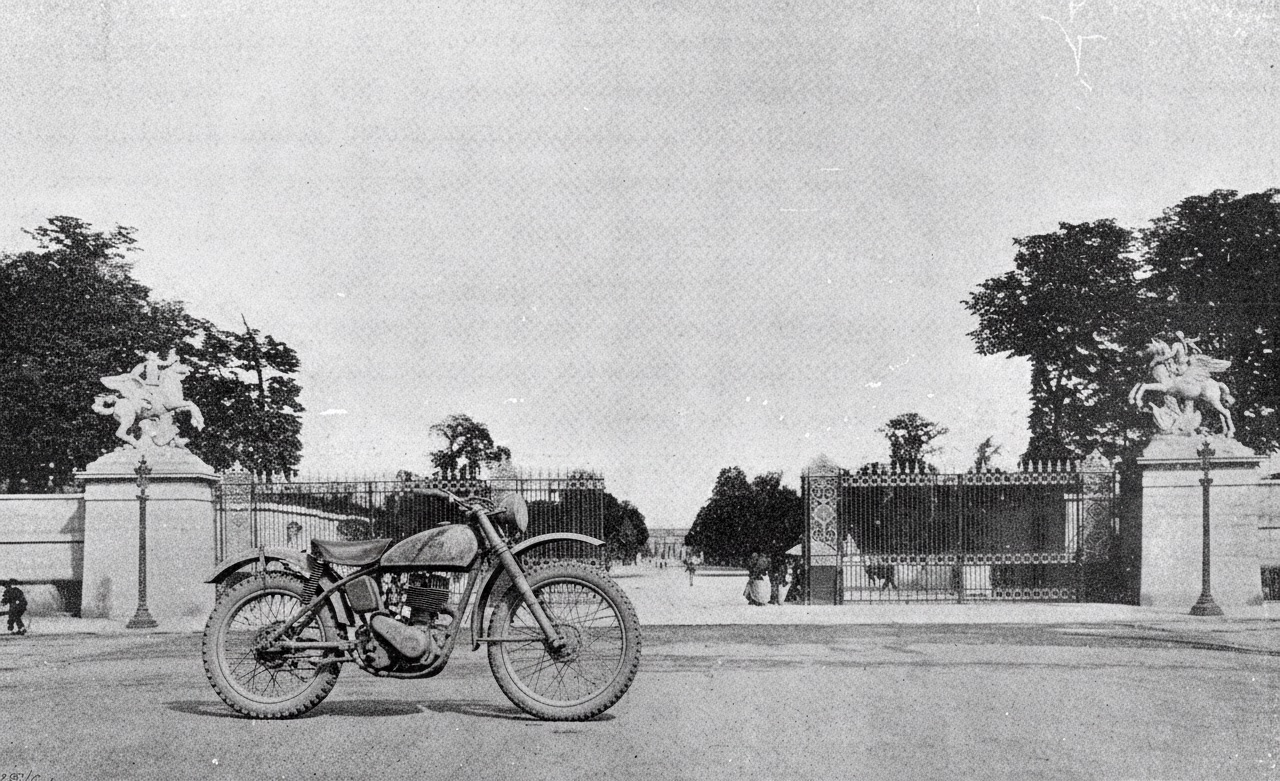} &
\includegraphics[width=0.305\textwidth]{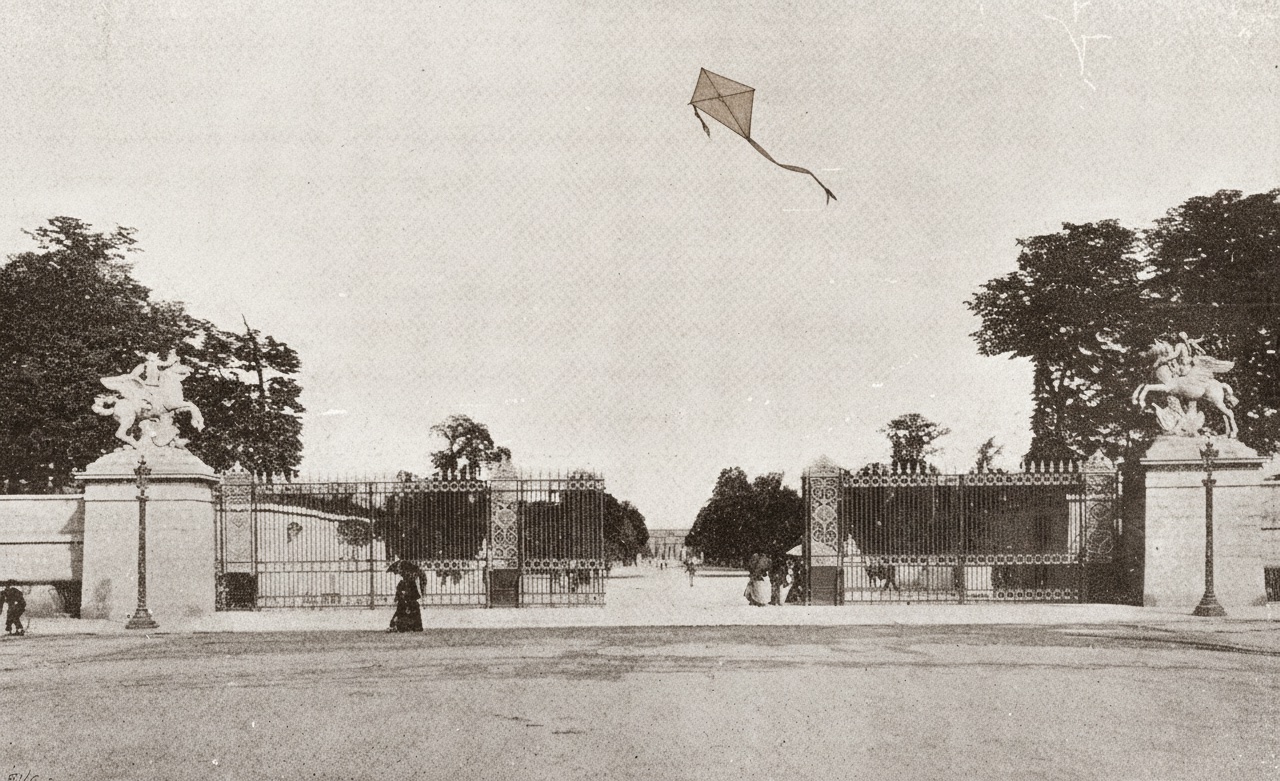} \\[-1pt]
& & \small\textit{Dirt Bike} & \small\textit{Kite} \\

\end{tabular}

\caption{Representative qualitative examples: each row shows the original image (left), an anachronistic image with a modern (post-1930) object (center), and a historically consistent image with an era-appropriate object (right); inserted objects are labeled below.}
\label{fig:qualitative}
\end{figure}

\begin{figure}[!ht]
\centering
\setlength{\tabcolsep}{2pt}
\renewcommand{\arraystretch}{1.0}
\small

\begin{tabular}{@{}rccc@{}}
& \textbf{Original}
& \textbf{Anachronistic}
& \textbf{Historically Consistent} \\[6pt]

\rotatebox{90}{\shortstack{RU\\(1913)}} &
\includegraphics[width=0.305\textwidth]{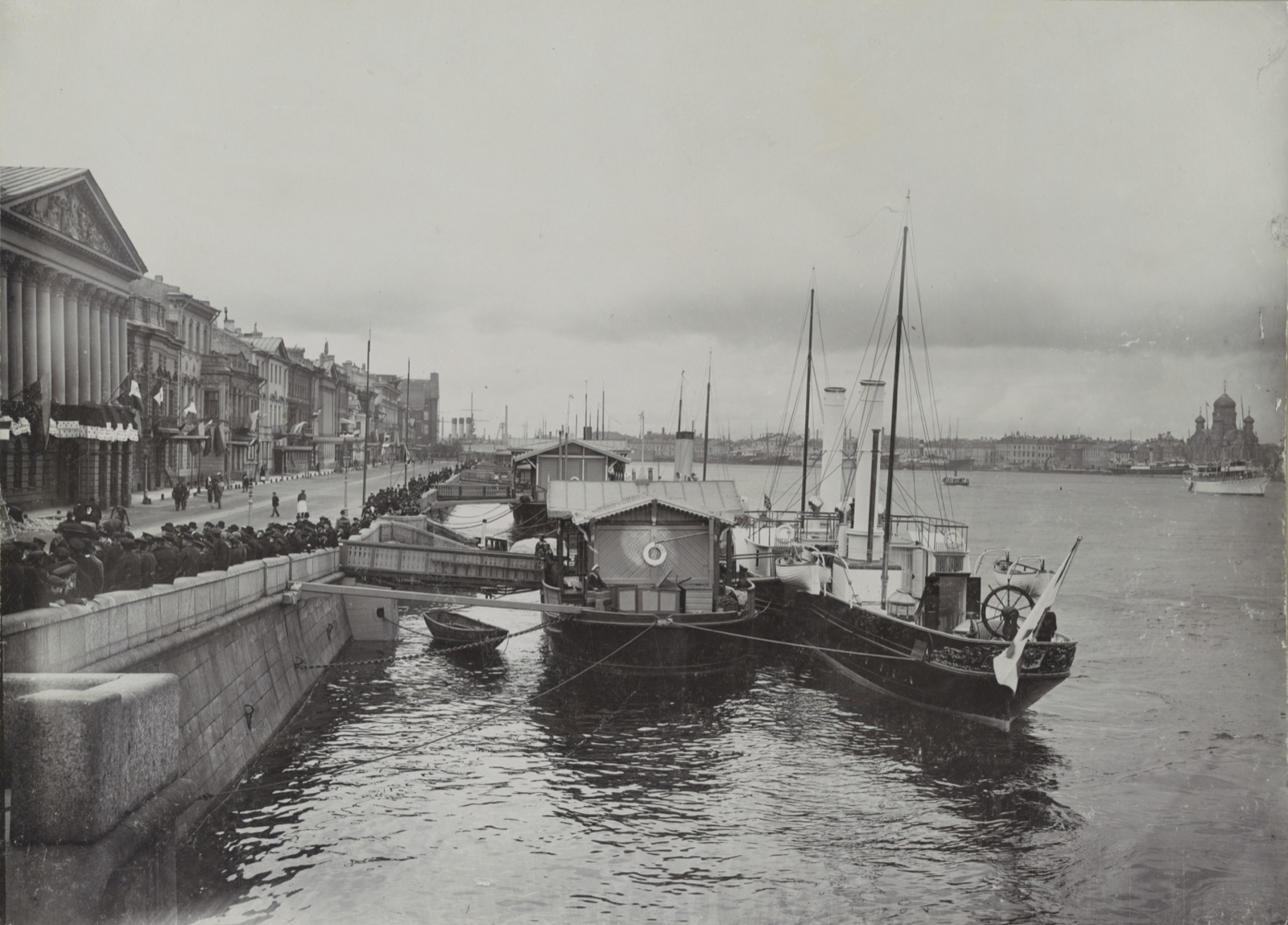} &
\includegraphics[width=0.305\textwidth]{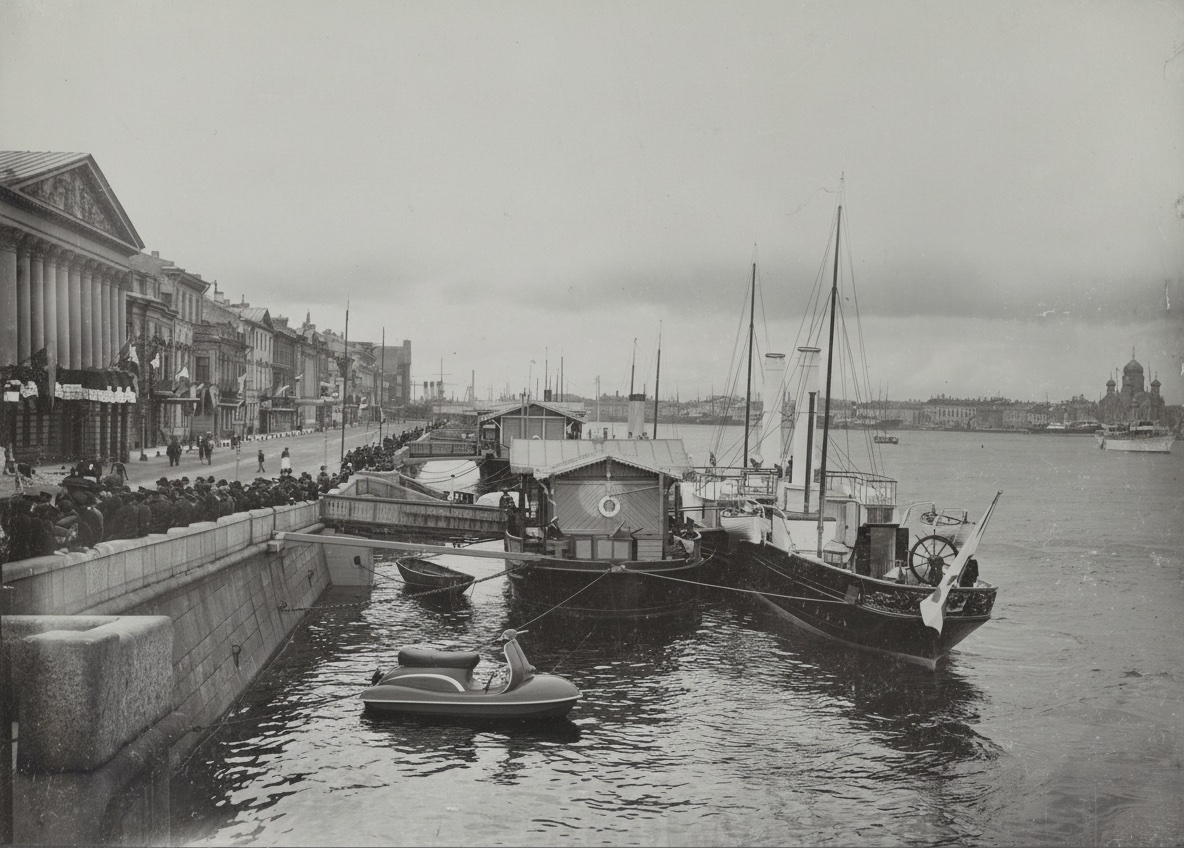} &
\includegraphics[width=0.305\textwidth]{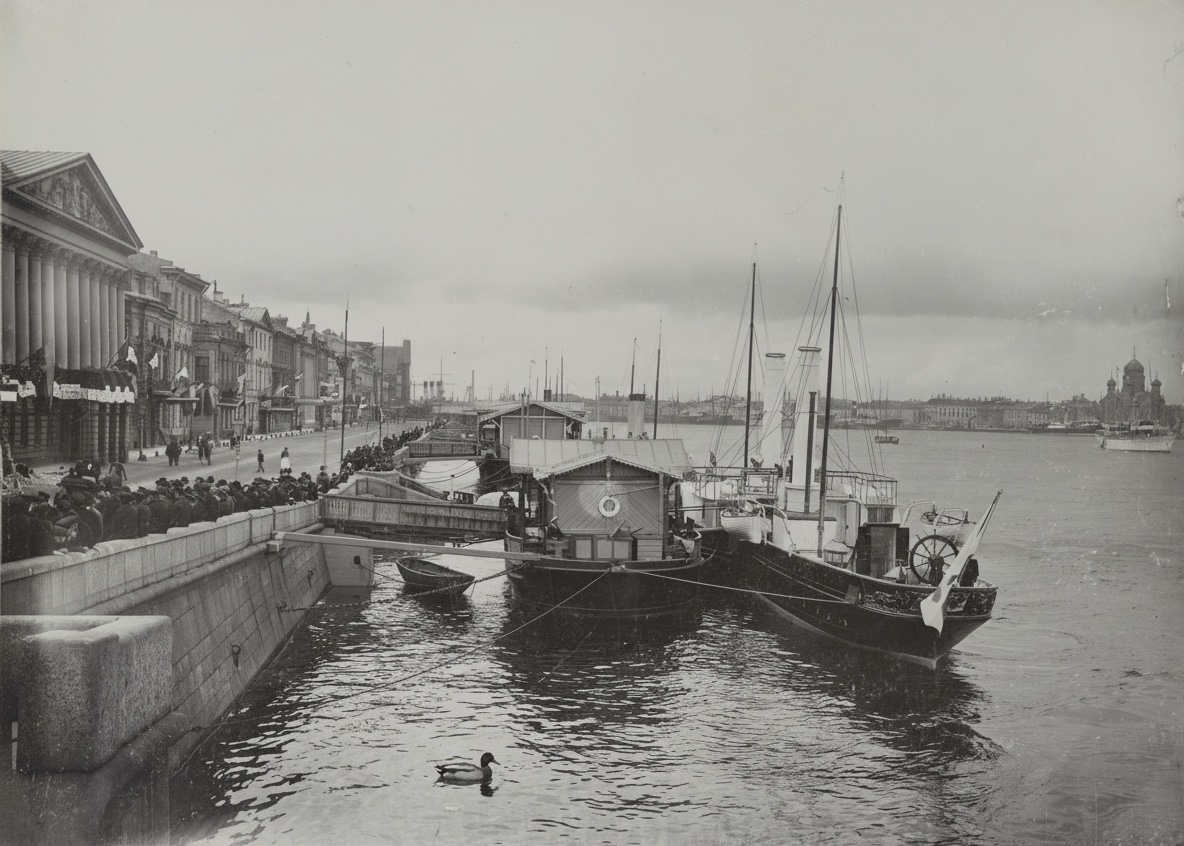} \\[-1pt]
& & \small\textit{Water Scooter} & \small\textit{Duck} \\[10pt]

\rotatebox{90}{\shortstack{CN\\(1909)}} &
\includegraphics[width=0.305\textwidth]{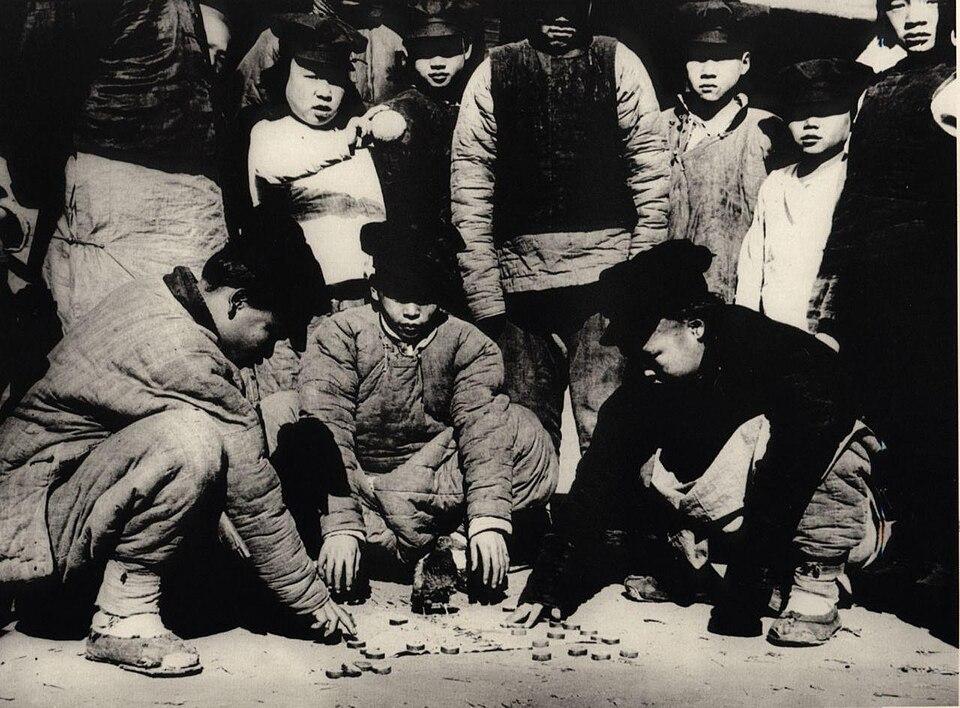} &
\includegraphics[width=0.305\textwidth]{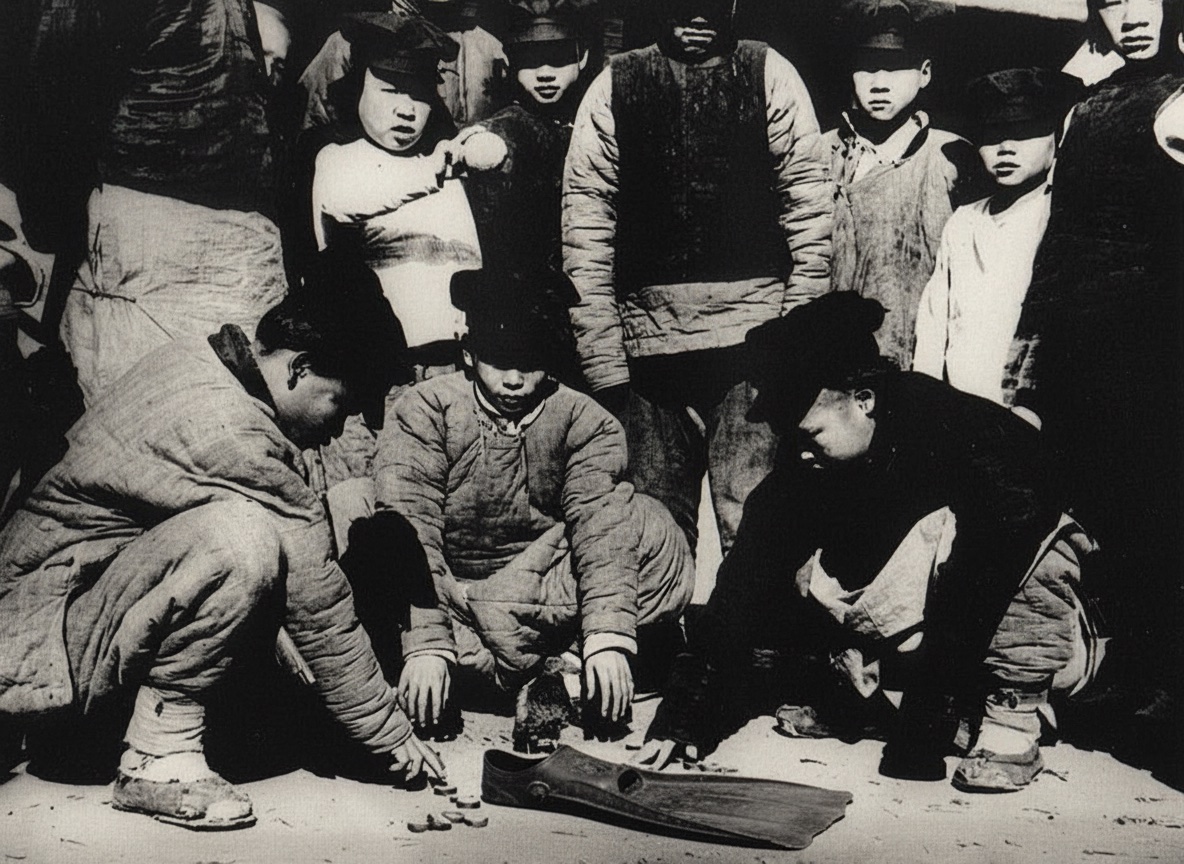} &
\includegraphics[width=0.305\textwidth]{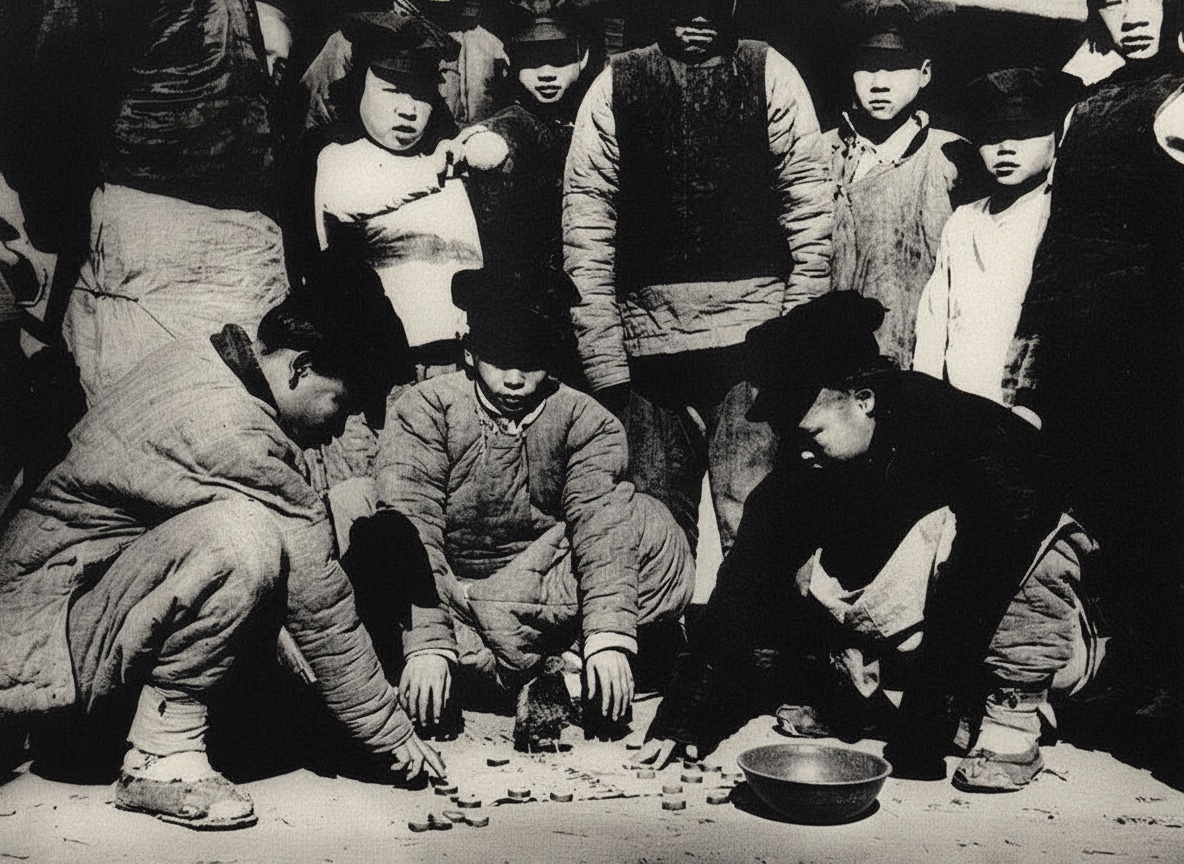} \\[-1pt]
& & \small\textit{Flipper} & \small\textit{Bowl} \\[10pt]

\rotatebox{90}{\shortstack{JP\\(1915)}} &
\includegraphics[width=0.305\textwidth]{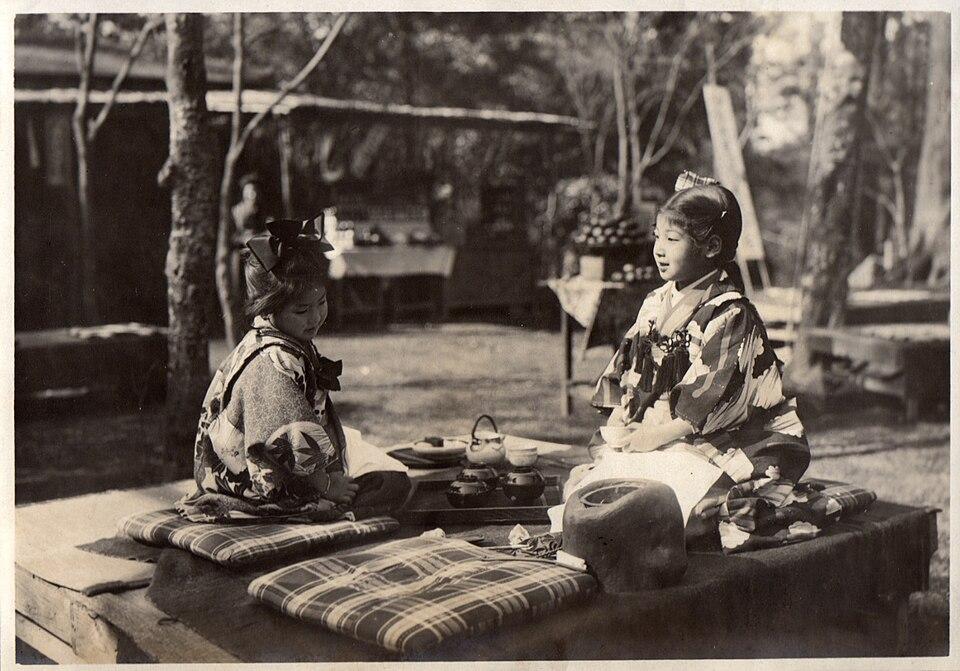} &
\includegraphics[width=0.305\textwidth]{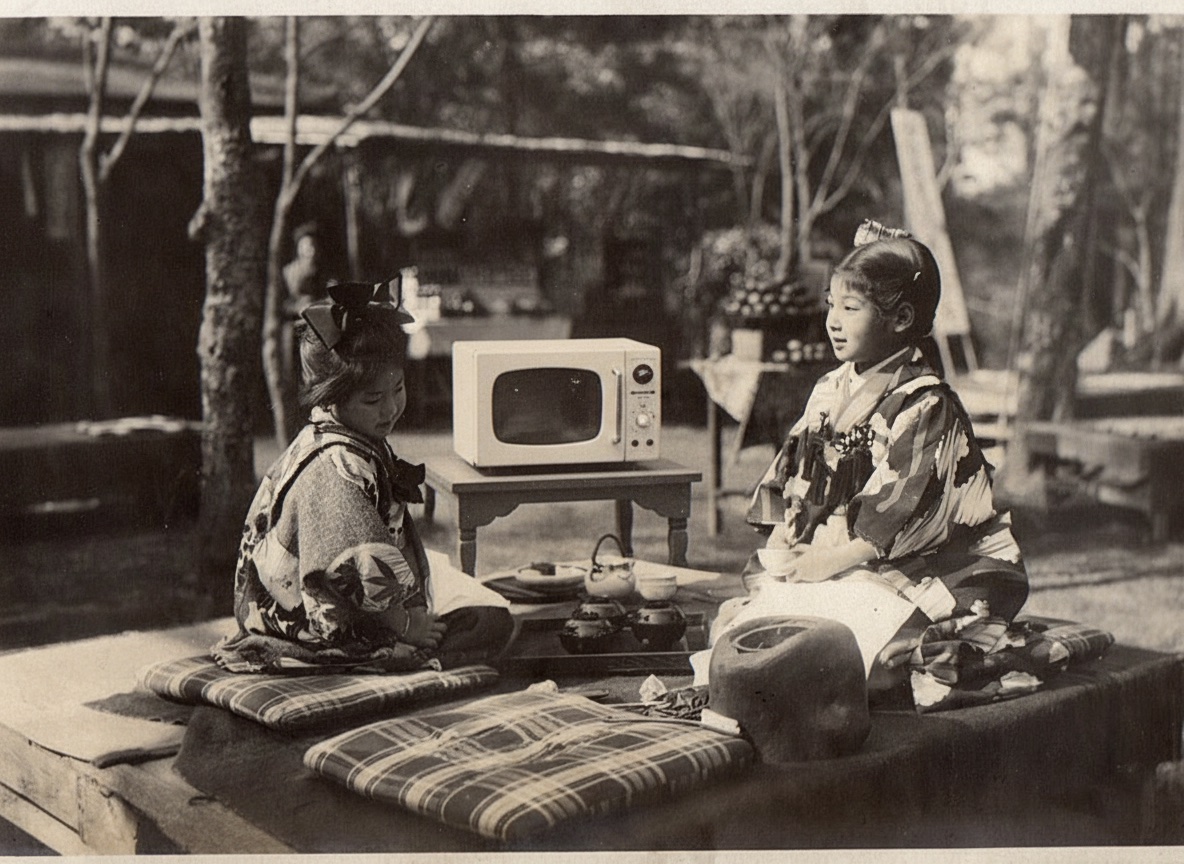} &
\includegraphics[width=0.305\textwidth]{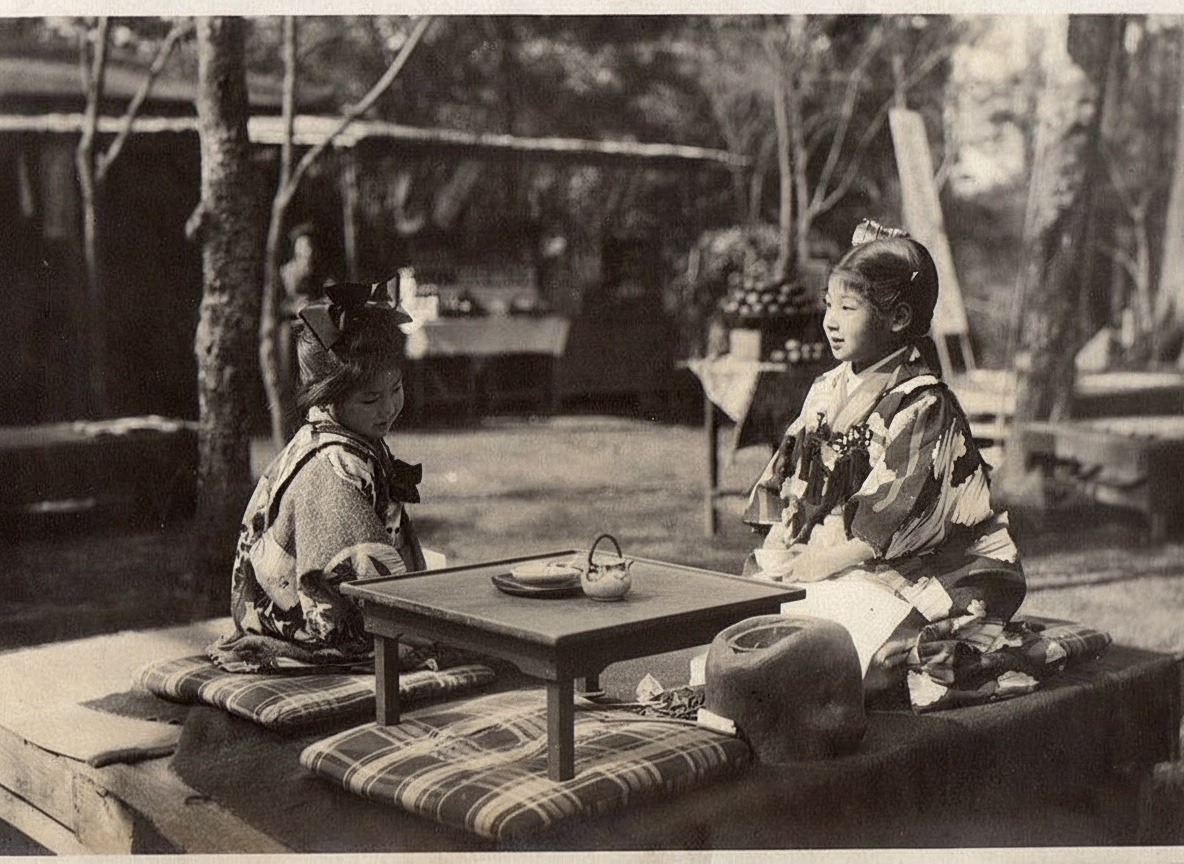} \\[-1pt]
& & \small\textit{Microwave Oven} & \small\textit{Table} \\[10pt]

\rotatebox{90}{\shortstack{KR\\(1920)}} &
\includegraphics[width=0.305\textwidth]{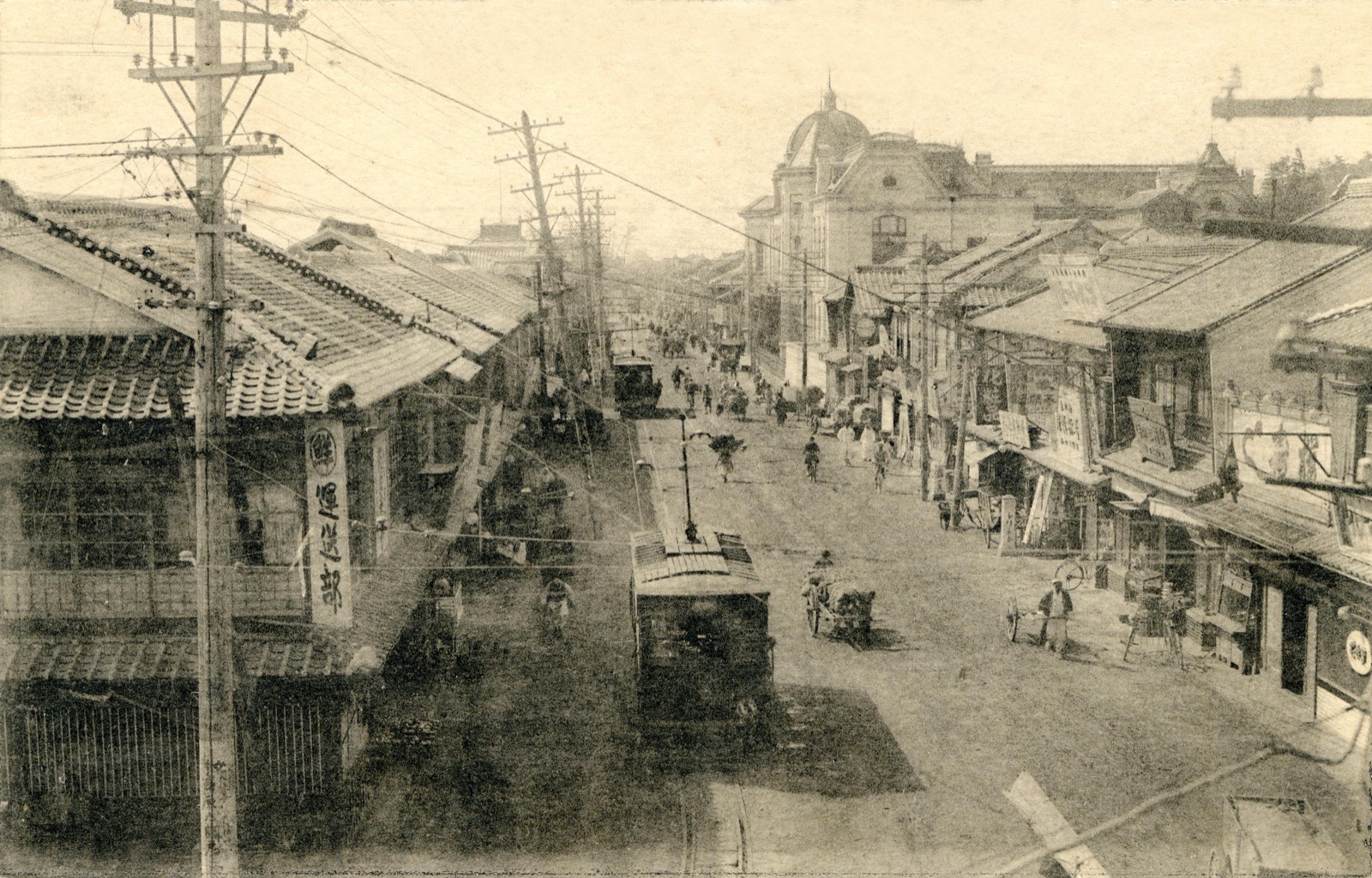} &
\includegraphics[width=0.305\textwidth]{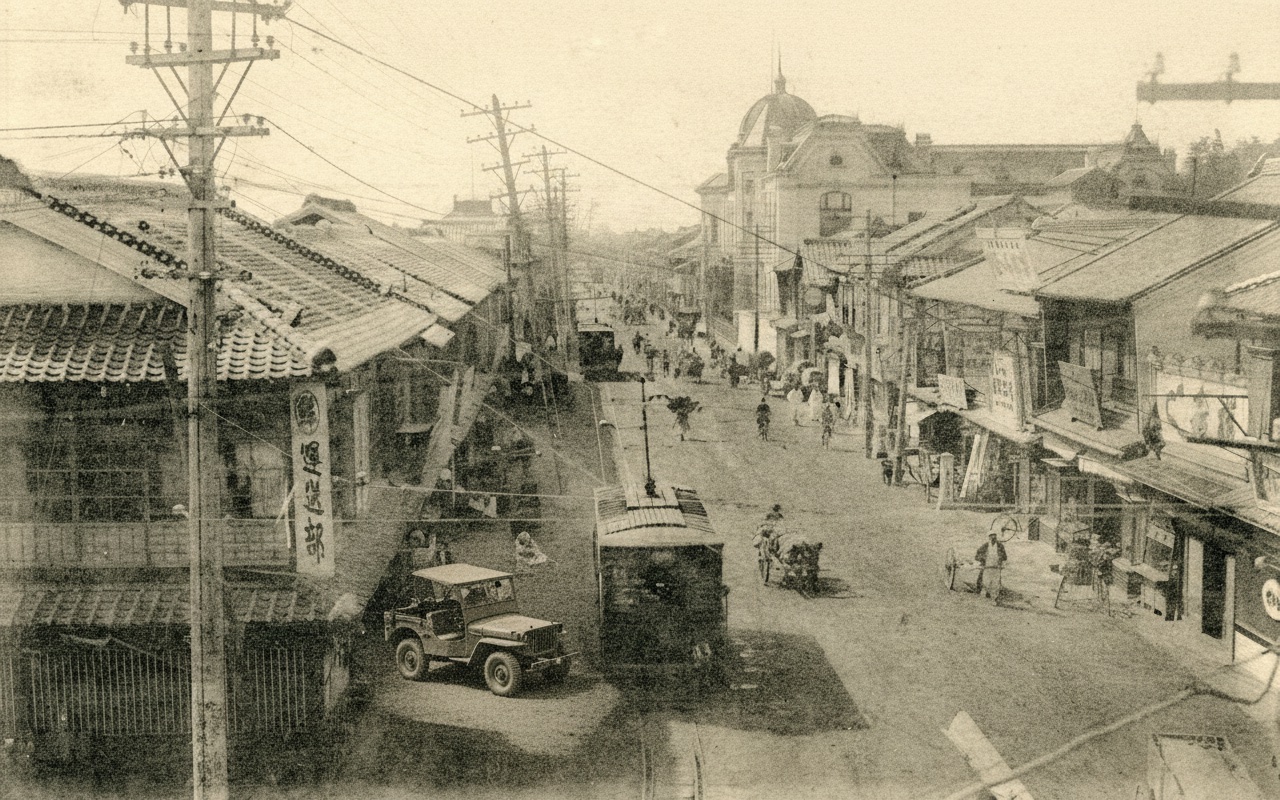} &
\includegraphics[width=0.305\textwidth]{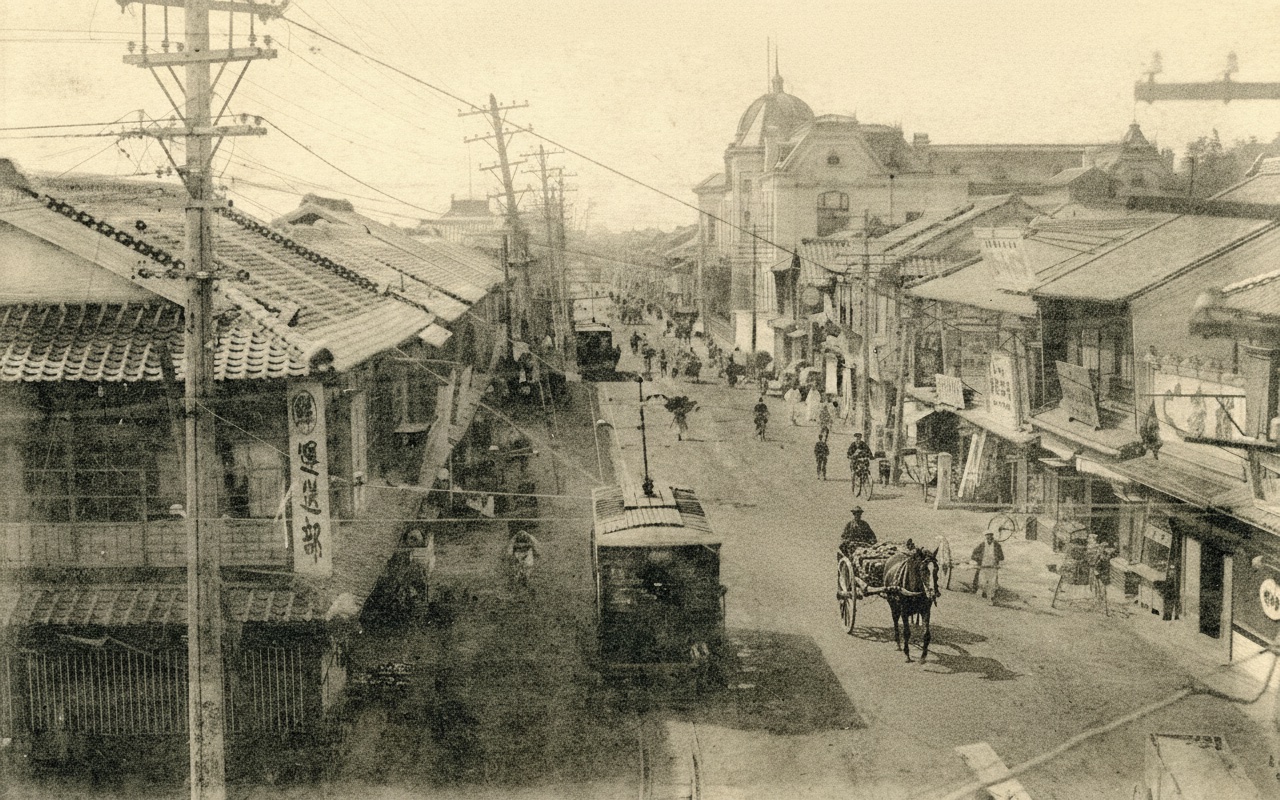} \\[-1pt]
& & \small\textit{Jeep} & \small\textit{Horse} \\

\end{tabular}

\caption{Additional representative examples showing original, anachronistic, and historically consistent images.}
\label{fig:qualitative2}
\end{figure}

Figures~\ref{fig:qualitative}--\ref{fig:qualitative2} show all three image types for each of the eight countries.

\clearpage

%% file: tables/benchmark_iaa.tex
\begin{table}[!htbp]
\caption{Inter-annotator agreement by country for anachronistic-image validation. The 3,489 candidates were reviewed before 1,600 images were selected for the benchmark.}
\label{tab:iaa}
\centering
 \footnotesize
\setlength{\tabcolsep}{3.5pt}
\renewcommand{\arraystretch}{1.0}
 \begin{tabular}{lccc}
\toprule
\textbf{Country} & \textbf{Candidates} & \textbf{Agreement} & \textbf{Cohen's $\kappa$} \\
\midrule
DE & 898 & 89.0\% & 0.779 \\
US & 522 & 84.9\% & 0.689 \\
NL & 481 & 89.2\% & 0.783 \\
FR & 434 & 95.6\% & 0.913 \\
RU & 465 & 82.8\% & 0.659 \\
KR & 444 & 97.7\% & 0.953 \\
CN & 136 & 77.2\% & 0.544 \\
JP & 109 & 89.0\% & 0.776 \\
\midrule
\textbf{Overall} & \textbf{3,489} & \textbf{89.1\%} & \textbf{0.781} \\
\bottomrule
\end{tabular}
\end{table}

%% file: tables/pre_filter_source.tex
\begin{table}[!htbp]
\setlength{\capinset}{0.05\linewidth}
\caption{Image distribution by source and country before filtering. Parenthetical ranges give the collection years for each country.}
\label{tab:pre_filter_source}
\centering
  \footnotesize
\setlength{\tabcolsep}{4pt}
\renewcommand{\arraystretch}{1.0}
\adjustbox{max width=\linewidth}{%
\begin{tabular}{lrrrrrrrrr}
\toprule
\textbf{Source}
  & \makecell[r]{\textbf{DE}\\\textbf{(1880--1919)}}
  & \makecell[r]{\textbf{US}\\\textbf{(1875--1919)}}
  & \makecell[r]{\textbf{NL}\\\textbf{(1880--1919)}}
  & \makecell[r]{\textbf{FR}\\\textbf{(1870--1919)}}
  & \makecell[r]{\textbf{RU}\\\textbf{(1880--1919)}}
  & \makecell[r]{\textbf{KR}\\\textbf{(1900--1929)}}
  & \makecell[r]{\textbf{CN}\\\textbf{(1880--1929)}}
  & \makecell[r]{\textbf{JP}\\\textbf{(1880--1929)}}
  & \textbf{Total} \\
\midrule
Wikimedia Commons & 3,677 & 695 & 691 & 529 & - & - & 153 & 121 & \textbf{5,866} \\
NYPL & - & 157 & - & - & - & - & - & - & \textbf{157} \\
Library of Congress & - & 331 & - & 184 & 878 & - & - & - & \textbf{1,393} \\
GongU Madang & - & - & - & - & - & 1,040 & - & - & \textbf{1,040} \\
\midrule
\textbf{Total} & \textbf{3,677} & \textbf{1,183} & \textbf{691} & \textbf{713} & \textbf{878} & \textbf{1,040} & \textbf{153} & \textbf{121} & \textbf{8,456} \\
\bottomrule
\end{tabular}%
}
\end{table}

%% file: tables/benchmark_source.tex
\begin{table}[!htbp]
\setlength{\capinset}{0.05\linewidth}
\caption{Image distribution by source and country after filtering and human validation. Parenthetical ranges give the retained image years for each country.}
\label{tab:source}
\centering
  \footnotesize
\setlength{\tabcolsep}{4pt}
\renewcommand{\arraystretch}{1.0}
\adjustbox{max width=\linewidth}{%
\begin{tabular}{lrrrrrrrrr}
\toprule
\textbf{Source}
  & \makecell[r]{\textbf{DE}\\\textbf{(1890--1919)}}
  & \makecell[r]{\textbf{US}\\\textbf{(1875--1919)}}
  & \makecell[r]{\textbf{NL}\\\textbf{(1890--1919)}}
  & \makecell[r]{\textbf{FR}\\\textbf{(1870--1919)}}
  & \makecell[r]{\textbf{RU}\\\textbf{(1880--1919)}}
  & \makecell[r]{\textbf{KR}\\\textbf{(1900--1928)}}
  & \makecell[r]{\textbf{CN}\\\textbf{(1880--1925)}}
  & \makecell[r]{\textbf{JP}\\\textbf{(1880--1929)}}
  & \textbf{Total} \\
\midrule
Wikimedia Commons & 441 & 112 & 232 & 128 & - & - & 49 & 40 & \textbf{1,002} \\
NYPL & - & 31 & - & - & - & - & - & - & \textbf{31} \\
Library of Congress & - & 123 & - & 82 & 185 & - & - & - & \textbf{390} \\
GongU Madang & - & - & - & - & - & 177 & - & - & \textbf{177} \\
\midrule
\textbf{Total} & \textbf{441} & \textbf{266} & \textbf{232} & \textbf{210} & \textbf{185} & \textbf{177} & \textbf{49} & \textbf{40} & \textbf{1,600} \\
\bottomrule
\end{tabular}%
}
\end{table}

%% file: tables/control_iaa.tex
\begin{table}[!htbp]
\caption{Inter-annotator agreement by country for historically consistent image validation. The 1,600 candidates were reviewed before 572 controls were selected.}
\label{tab:iaa_sub}
\centering
 \footnotesize
\setlength{\tabcolsep}{3.5pt}
\renewcommand{\arraystretch}{1.0}
 \begin{tabular}{lccc}
\toprule
\textbf{Country} & \textbf{Candidates} & \textbf{Agreement} & \textbf{Cohen's $\kappa$} \\
\midrule
DE & 441 & 85.9\% & 0.694 \\
US & 266 & 78.9\% & 0.576 \\
NL & 232 & 85.8\% & 0.708 \\
FR & 210 & 81.0\% & 0.622 \\
RU & 185 & 87.0\% & 0.731 \\
KR & 177 & 89.8\% & 0.773 \\
CN & 49 & 85.7\% & 0.692 \\
JP & 40 & 85.0\% & 0.694 \\
\midrule
\textbf{Overall} & \textbf{1,600} & \textbf{84.6\%} & \textbf{0.687} \\
\bottomrule
\end{tabular}
\end{table}

%% file: tables/gpt_compositing_full.tex
\begin{table}[!htbp]
\caption{Country-wise and overall conditional ECC (\%) for the 11 models rerun on 432 accepted \texttt{gpt-image-1.5} composites. Scores are computed on IOP-correct edited images, giving model-specific denominators; the smallest country cells contain about 20 images.}
\label{tab:gpt_compositing}
\centering
\footnotesize
\setlength{\tabcolsep}{5pt}
\renewcommand{\arraystretch}{1.0}

\adjustbox{max width=\linewidth}{%
\begin{tabular}{lccccccccc}
\toprule
\textbf{Model} & \textbf{DE} & \textbf{US} & \textbf{NL} & \textbf{FR} & \textbf{RU} & \textbf{KR} & \textbf{CN} & \textbf{JP} & \textbf{Overall} \\
\midrule
GPT-4o & 89.1 & 79.7 & 81.5 & 85.5 & 85.5 & 79.5 & 66.7 & 83.3 & 82.3 \\
Gemini-2.5-Flash & 96.1 & 96.3 & 87.0 & 95.1 & 96.4 & 98.6 & 90.5 & 92.0 & 94.6 \\
Gemini-2.5-Pro & 96.3 & 96.4 & 92.9 & 96.8 & 96.6 & 94.5 & 100.0 & 96.0 & 95.8 \\
\midrule
LLaVA-1.5-7B & 10.0 & 23.1 & 5.2 & 13.3 & 20.3 & 15.9 & 17.4 & 0.0 & 14.0 \\
CogVLM2-19B & 74.5 & 71.2 & 57.4 & 80.4 & 85.5 & 84.1 & 83.3 & 47.8 & 75.1 \\
Llama-3.2-11B & 26.8 & 42.4 & 12.9 & 44.6 & 28.8 & 23.0 & 32.0 & 12.0 & 28.7 \\
Gemma3-4B & 0.0 & 22.0 & 10.9 & 15.0 & 12.3 & 4.2 & 8.3 & 4.5 & 9.9 \\
Gemma3-12B & 55.4 & 54.5 & 45.6 & 73.0 & 70.7 & 63.9 & 50.0 & 41.7 & 59.2 \\
Gemma3-27B & 57.4 & 58.9 & 44.1 & 74.2 & 69.5 & 31.1 & 52.0 & 41.7 & 54.0 \\
InternVL3.5-8B & 60.7 & 71.2 & 44.8 & 74.2 & 66.7 & 22.2 & 29.2 & 13.6 & 51.7 \\
InternVL3.5-14B & 72.7 & 85.0 & 59.3 & 84.4 & 83.6 & 61.6 & 70.0 & 52.6 & 72.8 \\
\bottomrule
\end{tabular}%
}
\end{table}

%% file: tables/prompt_templates.tex
\begin{table}[!htbp]
\caption{ECC question templates used in the prompt-sensitivity analysis. The country and period placeholders are filled from benchmark metadata.}
\label{tab:prompt_templates}
\centering
\footnotesize
\setlength{\tabcolsep}{4pt}
\renewcommand{\arraystretch}{1.12}
\begin{tabular}{@{}lp{0.84\columnwidth}@{}}
\toprule
\textbf{Prompt} & \textbf{Template} \\
\midrule
Original & Given the stated setting of \texttt{\{country\}} during \texttt{\{period\}}, is this scene consistent with that time and place? \\
\cmidrule(lr){1-2}
Paraphrase 1 & Could this scene realistically have been photographed in \texttt{\{country\}} during \texttt{\{period\}}? \\
\cmidrule(lr){1-2}
Paraphrase 2 & Consider the historical context of \texttt{\{country\}} during \texttt{\{period\}}. Based only on what is visible, is the depicted scene historically consistent with that context? \\
\bottomrule
\end{tabular}
\end{table}

%% file: tables/aggregate_breakdowns.tex
\begin{table}[!htbp]
\caption{Country- and object-level performance (\%), computed per model then averaged over the 18 models. cECC is conditioned on correct object perception. Panel (b) gives the five highest- and five lowest-scoring objects among those with at least 30 pairs.}
\label{tab:aggregate_breakdowns}
\centering
\footnotesize
\setlength{\tabcolsep}{4.2pt}
\renewcommand{\arraystretch}{0.95}
\begin{minipage}[t]{0.48\linewidth}
\centering
\textbf{(a) Country-wise results}\\[3pt]
\begin{tabular*}{\linewidth}{@{\extracolsep{\fill}}lrrrr@{}}
\toprule
\textbf{Country} & \textbf{Pairs} & \textbf{IOP} & \textbf{ECC} & \textbf{cECC} \\
\midrule
DE & 441 & 75.3 & 30.4 & 35.3 \\
US & 266 & 77.0 & 34.4 & 44.8 \\
NL & 232 & 74.3 & 29.8 & 36.2 \\
FR & 210 & 78.1 & 34.1 & 41.3 \\
RU & 185 & 83.4 & 41.2 & 48.5 \\
KR & 177 & 84.4 & 32.5 & 39.1 \\
CN & 49 & 78.5 & 27.0 & 34.3 \\
JP & 40 & 78.9 & 28.9 & 34.7 \\
\midrule
Overall & 1,600 & 78.0 & 32.8 & 39.8 \\
\bottomrule
\end{tabular*}
\end{minipage}\hfill
\begin{minipage}[t]{0.48\linewidth}
\centering
\textbf{(b) Object-level results}\\[3pt]
\begin{tabular*}{\linewidth}{@{\extracolsep{\fill}}lrrrr@{}}
\toprule
\textbf{Object} & \textbf{Pairs} & \textbf{IOP} & \textbf{ECC} & \textbf{cECC} \\
\midrule
fighter jet & 86 & 89.5 & 58.4 & 62.9 \\
helicopter & 94 & 96.2 & 56.4 & 59.9 \\
drone & 55 & 89.8 & 47.5 & 52.9 \\
jet plane & 94 & 79.8 & 47.0 & 52.3 \\
minivan & 83 & 84.7 & 42.8 & 49.4 \\
\midrule
solar array & 90 & 72.5 & 21.9 & 27.6 \\
dish antenna & 93 & 74.7 & 18.8 & 23.5 \\
traffic cone & 34 & 92.8 & 17.2 & 20.1 \\
parking meter & 54 & 66.4 & 8.0 & 14.1 \\
frisbee & 32 & 49.5 & 5.4 & 10.3 \\
\bottomrule
\end{tabular*}
\end{minipage}
\end{table}